\documentclass[letterpaper,11pt]{article}

\usepackage{diagbox}
\usepackage{makecell}
\usepackage{booktabs}
\usepackage{tikz,tikz-3dplot}
\usepackage{bbm}
\usepackage{amsfonts}
\usepackage{amsmath}
\usepackage{amssymb}
\usepackage{amsthm}
\usepackage{url,ifthen}
\usepackage[shortlabels]{enumitem}
\usepackage{srcltx}
\usepackage{dsfont}
\usepackage{multirow}
\usepackage{boxedminipage}
\usepackage[margin=1.1in]{geometry}
\usepackage{nicefrac}
\usepackage{xspace}
\usepackage{graphicx}
\usepackage{color}
\usepackage{subfigure}
\usepackage{colortbl}
\usepackage{setspace}
\usepackage{pgfplots}
\pgfplotsset{compat=1.12}
\usepackage{algorithm,algorithmic}
\usepackage[algo2e]{algorithm2e}

\usepackage{xcolor}
\definecolor{DarkGreen}{rgb}{0.1,0.5,0.1}
\definecolor{DarkRed}{rgb}{0.5,0.1,0.1}
\definecolor{DarkBlue}{rgb}{0.1,0.1,0.5}
\definecolor{Gray}{rgb}{0.2,0.2,0.2}

\definecolor{c1}{RGB}{38, 70, 83}
\definecolor{c2}{RGB}{42, 157, 143}
\definecolor{c3}{RGB}{233, 196, 106}
\definecolor{c5}{RGB}{231, 111, 81}
\definecolor{c4}{RGB}{244, 162, 97}

\definecolor{base}{rgb}{0.23299120924703914, 0.639586552066035, 0.9260706093977744}
\definecolor{instruct}{rgb}{0.21044753832183283, 0.6773105080456748, 0.6433941168468681}
\definecolor{instruct}{RGB}{229, 128, 145}
\definecolor{census}{rgb}{0.3126890019504329, 0.6928754610296064, 0.1923704830330379}
\definecolor{uniform}{rgb}{0.9333333333333333, 0.5215686274509804, 0.2901960784313726}

\usepackage{listings}
\lstdefinestyle{mystyle}{
    commentstyle=\color{DarkBlue},
    keywordstyle=\color{DarkRed},
    numberstyle=\tiny\color{Gray},
    stringstyle=\color{DarkGreen},
    basicstyle=\footnotesize,
    breakatwhitespace=false,         
    breaklines=true,                 
    captionpos=b,                    
    keepspaces=true,                 
    numbers=left,                    
    numbersep=5pt,                  
    showspaces=false,                
    showstringspaces=false,
    showtabs=false,                  
    tabsize=2
}
\usepackage[small]{caption}

\usepackage[pdftex]{hyperref}
\hypersetup{
    unicode=false,          
    pdftoolbar=true,        
    pdfmenubar=true,        
    pdffitwindow=false,      
    pdfnewwindow=true,      
    colorlinks=true,       
    linkcolor=DarkBlue,          
    citecolor=DarkGreen,        
    filecolor=DarkRed,      
    urlcolor=DarkBlue,          
    pdftitle={},
    pdfauthor={},
}

\def\draft{1}

\def\submit{0}

\ifnum\draft=1 
    
\else
    
\fi

\ifnum\submit=1
\newcommand{\forsubmit}[1]{#1}
\newcommand{\forreals}[1]{}
\else
\newcommand{\forreals}[1]{#1}
\newcommand{\forsubmit}[1]{}
\fi

\newtheorem*{definition*}{Definition}

\theoremstyle{definition}

\newtheoremstyle{example_contd}
{\topsep} {\topsep}%
{}
{}
{\bfseries}
{.}
{1em}
{\thmname{#1} \thmnumber{ #2}\thmnote{#3} (continued)}

\theoremstyle{example_contd}

\newcommand{\chapterref}[1]{\hyperref[ch:#1]{Chapter~\ref{ch:#1}}}

\newcommand{\claimref}[1]{\hyperref[claim:#1]{Claim~\ref{claim:#1}}}

\newcommand{\corollaryref}[1]{\hyperref[cor:#1]{Corollary~\ref{cor:#1}}}

\newcommand{\definitionref}[1]{\hyperref[def:#1]{Definition~\ref{def:#1}}}

\newcommand{\equationref}[1]{\hyperref[eq:#1]{Equation~\ref{eq:#1}}}

\newcommand{\factref}[1]{\hyperref[fact:#1]{Fact~\ref{fact:#1}}}

\newcommand{\figureref}[1]{\hyperref[fig:#1]{Figure~\ref{fig:#1}}}

\newcommand{\tableref}[1]{\hyperref[tab:#1]{Table~\ref{tab:#1}}}

\newcommand{\itemref}[1]{\hyperref[item:#1]{Item~(\ref{item:#1})}}

\newcommand{\lemmaref}[1]{\hyperref[lem:#1]{Lemma~\ref{lem:#1}}}

\newcommand{\propref}[1]{\hyperref[prop:#1]{Proposition~\ref{prop:#1}}}

\newcommand{\propositionref}[1]{\hyperref[prop:#1]{Proposition~\ref{prop:#1}}}

\newcommand{\remarkref}[1]{\hyperref[rem:#1]{Remark~\ref{rem:#1}}}

\newcommand{\sectionref}[1]{\hyperref[sec:#1]{Section~\ref{sec:#1}}}

\newcommand{\theoremref}[1]{\hyperref[thm:#1]{Theorem~\ref{thm:#1}}}

\usepackage[charter]{mathdesign}

\usepackage{microtype}

\newcommand{\cF}{{\cal F}}

\renewcommand{\geq}{\geqslant}

\newcommand{\C}{\mathbb C}

\usepackage{bm}

\newcommand{\remove}[1]{}

\usepackage{fontawesome}

\usepackage{caption}
\usepackage{xcolor}
\usepackage{xurl}
\usepackage{tabularx} 

\definecolor{myblue}{rgb}{0.23299120924703914, 0.639586552066035, 0.9260706093977744}
\definecolor{myred}{rgb}{0.9677975592919913, 0.44127456009157356, 0.5358103155058701}
\definecolor{mygreen}{rgb}{0.3126890019504329, 0.6928754610296064, 0.1923704830330379}
\definecolor{myorange}{rgb}{0.9333333333333333, 0.5215686274509804, 0.2901960784313726}

\usepackage{xspace}
\usepackage[bottom]{footmisc}
\usepackage{soul}

\usepackage[square]{natbib}

\usepackage{pifont}

\title{Questioning the Survey Responses of Large Language Models} 

\date{ $^*$\textit{Max Planck Institute for Intelligent Systems, T\"ubingen, Germany\\ $^\S$ELLIS Institute Tübingen, Germany\\$^\ddagger$T\"ubingen AI Center}
}
\author{Ricardo Dominguez-Olmedo$^{*\ddagger}$ \and  Moritz Hardt$^{*\ddagger}$\and Celestine Mendler-D\"unner$^{*\S\ddagger}$}

\begin{document}

\maketitle

\begin{abstract}
Surveys have recently gained popularity as a tool to study large language models. By comparing survey responses of models to those of human reference populations, researchers aim to infer the demographics, political opinions, or values best represented by current language models. In this work, we critically examine this methodology on the basis of the well-established American Community Survey by the U.S. Census Bureau. Evaluating 43 different language models using de-facto standard prompting methodologies, we establish two dominant patterns. First, models' responses are governed by ordering and labeling biases, for example, towards survey responses labeled with the letter `A'.
Second, when adjusting for these systematic biases through randomized answer ordering, models across the board trend towards uniformly random survey responses, irrespective of model size or pre-training data. As a result, in contrast to conjectures from prior work, survey-derived alignment measures often permit a simple explanation: models consistently appear to better represent subgroups whose aggregate statistics are closest to uniform for any survey under consideration.
\end{abstract}


\section{Introduction}

Surveys have a long tradition in social science research as a means for gathering statistical information about the characteristics, values, and opinions of human populations~\citep{groves2009survey}.
Insights from surveys inform policy interventions, business decisions, and science across various domains. 
Surveys typically consist of a series of well-curated questions in a multiple-choice format, with unambiguous framing and a set of answer choices carefully selected by domain experts.
Surveys are then presented to groups of individuals and their answers are aggregated to gain statistical insights about the populations that these groups of individuals represent.

Many established survey questionnaires together with the carefully collected answer statistics are publicly available.
Machine learning researchers have identified the potential benefits of building on this valuable data resource to study large language models (LLMs). Survey questions offer a way to systematically prompt LLMs, and the aggregate statistics over answers collected by surveying human populations serve as a reference point for evaluation. As a result, the use of surveys has recently gained popularity for studying LLMs' biases \citep{santurkar2023opinions, durmus2023towards}. 
Also prompting LLMs with survey questions, researchers in the social sciences have explored using LLMs to emulate the survey responses of human populations~\citep{argyle2022out, lee2023can}. 
If effective proxies, simulated responses could 
augment or replace the expensive data collection process involving human subjects and provide insights into subpopulations that are otherwise hard to reach.

It is tempting to prompt LLMs with survey questions, due to their syntactic similarity to question answering tasks~\citep{brown20gpt3, liang2022holistic}. However, it is a priori unclear how to interpret their answers. Rather than knowledge testing, surveys seek to elicit aggregate statistics over individuals, providing an unbiased view on the properties of the population they are targeting. The quality of survey data hinges on the validity and robustness of the conclusions that can be drawn from it. 
Clearly, running a survey on LLMs is different from interrogating humans and thus it comes with distinct challenges.
While much research has gone into carefully designing surveys to ensure faithful human responses, it is unclear whether prompting LLMs with the same surveys satisfies similar premises. We devote this work to gain systematic insights into the survey responses of LLMs, what we can expect to learn from them, and to what extent they resemble those of human populations.

\subsection{Our work}

\begin{figure}[t!]
\centering
\includegraphics[width=\linewidth]{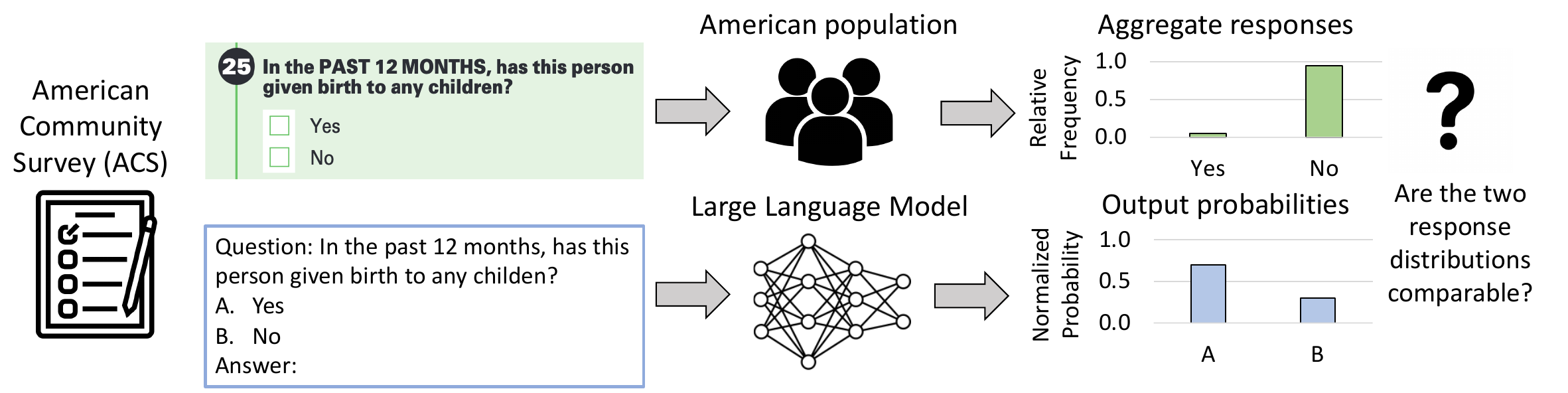}
\caption{We prompt language models with questions from the American Community Survey (ACS). We systematically compare models' survey responses to those of the U.S. Census.}
\label{fig:teaser}
\end{figure}

The basis of our investigation is the American Community Survey\footnote{\href{https://www.census.gov/programs-surveys/acs}{https://www.census.gov/programs-surveys/acs}}~(ACS), a demographic survey conducted by the U.S. Census Bureau at a national level, on a yearly basis. We curate a questionnaire composing of 25 multiple choice questions from the 2019 ACS. We prompt 43 language models of varying size with these questions, individually and in sequence, and we record their probability distribution over answers. Based on the collected data, we investigate the following two questions:
\emph{What can we infer about LLMs, and the data they have been trained on, from their survey responses? Does the data generated by prompting models to answer the ACS questionnaire qualitatively resemble the census data collected by surveying the U.S. population?} See Figure~\ref{fig:teaser}.

We start by inspecting models' distributions over answers to individual survey questions when the questions are asked independently. We observe that the entropy of response distributions differs substantially across models of varying size. Entropy tends to increase log linearly with model size, and it is preserved across different questions asked. We find that this differences arise because strong ordering and labeling biases confound models' answers. In fact, after adjusting for such systematic biases through randomized choice ordering, we find that response distributions are very similar across models and tend to correspond to highly balanced answers.

Comparing models' adjusted responses to those of the U.S. census population, we find that natural variations in entropy across questions are not reflected in the responses. Instead, on average across questions,  models' responses are no closer to the census population, or the population of any state within the US, than to a fixed uniform baseline. This qualitative difference between model responses and human data puts into question the insights that can be gained from such comparisons. We find that even after instruction-tuning this trend persists, and model responses have consistently higher entropy than any human population we compare to, independent of the survey used. Only for models of size larger than $70$ billion parameters we can recognize a trend that the divergence between model responses and the census data decreases after instruction-tuning.

With these insights in mind, we inspect conjectures from prior work related to survey derived alignment metrics, that is, that differences in similarity between models' and populations' responses might be attributable to certain demographics being better represented in the training data. Instead, our results suggest a much simpler explanation: the relative alignment of model responses with different demographic subgroups can be explained by the entropy of the subgroups' responses, irrespective of the data or training procedure employed to train the model. We demonstrate this beyond the ACS on other surveys considered by prior work. As such, our findings provide important context to prior studies that employ surveys to examine the biases of LLMs.

More broadly, our findings suggest caution when treating language models' survey responses as a faithful representation of any human population, at least a present time, as it could lead to potentially misguided conclusions about alignment.

\subsection{Related work}

Despite the syntactical similarities, there are important differences between  evaluating LLMs on the basis of their survey responses and traditional question answering evaluations~\citep{liang2022holistic}.
Question answering (QA) tasks predominantly serve the purpose of knowledge testing~\citep[e.g.,][]{kwiatkowski2019natural, rajpurkar2016squad, talmor2019ommonsenseqa, mihaylov2018can}. In such setting, a language model's answer to some unambiguous input question is extracted by computing its most likely completion. Similarly, for questions that lack a clear answer (e.g., ``Angela and Patrick are sitting together. Who is an entrepreneur?'') models' most likely response have been used to investigate various biases of LLMs~\citep{li2020unqovering, mao2021eliciting, perez2022discovering, abid2021persistent, jiang2022communitylm}. 

When evaluating LLMs on the basis of survey questions, the focus is not on the model’s most likely completion but rather on the probability distribution that the model assigns to various answer choices. For example, not whether the model is more likely to answer ``Yes'' than ``No'' to a given survey question, but the normalized probability assigned to each of the two answer choices. See Figure~\ref{fig:teaser}.
More concretely, \citet{santurkar2023opinions} study LLMs' answer distributions for multiple-choice opinion polling questions, measuring their similarity to those of various U.S. demographic groups. They extract models' answer distributions from the next token probabilities corresponding to each answer choice. Subsequent works employ a similar methodology but instead consider transnational opinion surveys~\citep{durmus2023towards, alkhamissi2024investigating} and moral beliefs surveys~\citep{scherrer2024evaluating}.
We adopt this popular methodology to systematically investigate the properties of models’ answer distributions on the basis of a well-established demographic survey.

Instead of asking questions individually, \citet{hartmann2023political, rutinowski2023self, motoki2023more, feng2023pretraining} sequentially prompt language models to answer entire political compass or voting advice questionnaires. Rather than aggregating answers into a political affinity score, our focus is instead on examining whether models' responses qualitatively resemble those of human populations. We discuss this sequential generation setting in detail in Appendix~\ref{sec:generation}.

Lastly, there is an emerging body of research that integrates LLMs into computational social science~\citep{ziems2023large}. 
This includes tasks such as taxonomic labeling, where language models are employed for tasks such as opinion prediction~\citep{kim2023aiaugmented, mellon2022do}, and free-form coding, where language models are used to generate explanations for social science constructs~\citep{nelson2021future}. 
Recent studies have also investigated the feasibility of using LLMs to simulate human participants in psychological, psycholinguistic, and social psychology experiments~\citep{DILLION2023, aher2023using}, or as proxies for specific human populations in social science research~\citep{argyle2022out, lee2023can, sanders2023demonstrations} and economics~\citep{brand2023market, horton2023large}. Within this context, our work suggests caution in relying on the survey responses of LLMs  to elicit synthetic responses that resemble those of human populations and highlights potential pitfalls.


\section{Surveying language models}\label{sec:sec2}

We employ the de-facto standard methodology to survey language models introduced by \citet{santurkar2023opinions}. For every survey question, we generate a prompt containing the multiple-choice question and we collect language models' probability distribution over answer choices. Formally, for a given model $m$ and survey question $q$ we define the model's \emph{survey response} as a categorical random variable $R_q^m$ which can take on $k_q$ values corresponding to the number of answer choices to question $q$. The respective answer distributions are then contrasted with those of human populations align various dimensions. The overall setup is illustrated in Figure~\ref{fig:teaser}. 

\paragraph{Prompting.}
We determine the event probabilities of $R_q^m$ by prompting model $m$ as follows:
\begin{enumerate}[itemsep=0pt,leftmargin=20pt]
    \item We construct an input prompt of the form 
    ``\texttt{\small Question: <question> \textbackslash n A. <choice 1 \allowbreak>\textbackslash n B. <choice 2> \textbackslash n  ... <choice $k_q$> \textbackslash n Answer:}''. 
    \item We query language models with the input prompt and obtain their output distribution over next-token probabilities. We select the $k_q$ output probabilities corresponding to each answer choice (e.g., the tokens ``A'', ``B'', etc.), and we renormalize to obtain the probability distribution over survey answers. \footnote{For OpenAI's models, we only have access to the top-5 next-token log probabilities through the OpenAI API. In this case, we assign to the unseen probabilities (if any) the minimum between the remaining probability mass and the smallest observed probability, following the methodology of \citet{santurkar2023opinions}}.
\end{enumerate}

The chosen style of prompt is standard for question answering tasks~\citep{hendrycks2021measuring}, used in OpinionQA~\citep{santurkar2023opinions}, and follows the best practices for social science research recommended by \citet{ziems2023large}. 
For completeness we perform several prompt ablations, including the prompt variations used by \citet{argyle2022out}, \citet{santurkar2023opinions} and \citet{durmus2023towards}. We find our take-aways to be robust to such changes, see Appendix~\ref{sec:ablation-prompt}. However, note that our goal is not to engineer better prompts, but to critically examine popular scientific practices. 

\paragraph{Survey questions.} We use a representative subset of 25 multiple-choice questions from the 2019 ACS questionnaire. We denote the set of questions by $Q$.
The questions cover basic demographic information, education attainment, healthcare coverage, disability status, family status, veteran status, employment status, and income. We generally consider the questions and answers as they appear in the ACS questionnaire. Figure~\ref{fig:teaser} depicts an example question. We refer to Appendix~\ref{sec:questionnaire} for our list of questions and the exact framing we used for each question.

\paragraph{Models surveyed.} We survey 43 language models of size varying from 110M to 175B parameters: the base models GPT-2~\citep{radford2019language}, GPT-Neo~\citep{gptneo}, Pythia~\citep{biderman2023pythia}, MPT~\citep{mpt}, Llama 2~\citep{touvron2023llama2},Llama~3~\citep{dubey2024llama3} and GPT-3  \citep{brown20gpt3}; as well as the instruct variants of MPT 7B and GPT NeoX 20B, the Dolly fine-tune of Pythia 12B~\citep{dolly}, Llama 2 Chat, Llama 3 Instruct, the text-davinci variants of GPT-3~\citep{rlhf}, and GPT-4~\citep{gpt42023}.

\paragraph{Reference data \& evaluation.} We use the responses collected by the U.S. Census Bureau when surveying the U.S. population as our reference data. In particular, we use the 2019 ACS public use microdata sample\footnote{\href{https://www.census.gov/programs-surveys/acs/microdata.html}{https://www.census.gov/programs-surveys/acs/microdata}} (henceforth census data). The data contains the anonymized responses of around 3.2 million individuals in the United States. For each survey question $q\in Q$, we denote the census' population-level response as a categorical random variable $C_q$ whose event probabilities are the relative frequency of each answer choice among survey respondents. We use $U_q$ to denote the uniform distribution over answers. Given these two reference points, we evaluate language models' responses $R_q^m$ along two dimensions:

\begin{itemize}[leftmargin=20pt]
    \item We use \emph{entropy} to measure the degree of variation in models' responses. We denote the entropy of a random variable $R$ as $H(R)$. To meaningfully compare the entropy of responses across questions with varying number of choices $k_q$, we report normalized entropy, that is, the entropy relative to the uniform distribution. 
$H(R_q^m)=1$ implies that model $m$'s survey response to question $q$ is uniformly distributed (i.e., $H(U_q)=1$). 
    \item We use the \emph{Kullback–Leibler (KL) divergence} to measure the ``similarity'' between two distributions over answers. We write $\mathrm{KL}(R_q^m \parallel C_q)$ for the KL divergence between the response distribution $R_q^m$ of model $m$ to question $q$ and the corresponding aggregate response distribution $C_q$ observed in the census data. The larger the KL distance between two distributions, the more dissimilar the two distributions are. 
\end{itemize}

Note that the KL divergence between any distribution and the uniform distribution corresponds to the entropy difference. 
For normalized entropy this yields $\mathrm{KL}(C_q\parallel U_q)=k_q(1-H(C_q))$.

\paragraph{Randomized choice ordering.} For several investigations we survey models under randomized choice ordering. This means, for a given question $q$, we prompt models with different permutations of the answer choice ordering, i.e., the assignment of answers (e.g., ``male'', ``female'') to choice labels (``A'', ``B'', etc), while the choice labels are kept in alphabetic order.
We evaluate models' survey responses under all possible choice orderings and we use $\bar R_q^{m}$ to denote the expected distribution over answers and $\bar O_q^m$ to denote the expected distribution over selected choice labels. For questions with more than  6 answers we evaluate a maximum of 5000 permutations. For OpenAI's models we evaluate up to 50 permutations due to the costs of querying the OpenAI API.
This distinction serves to decouple a model's tendency towards picking a particular answer from its tendency towards picking a particular choice label.
In the following we refer to the expected survey response $\bar R^{m}_q$ under uniformly distributed choice ordering as the \emph{adjusted} survey response. We will come back to this in Section~\ref{sec:adjusted}.~\looseness=-1

\section{Systematic biases in models' survey responses}
\label{sec:initial-obs}

We start by surveying the base pre-trained models.
We present survey questions independently of one another, showing the answer choices in the same order as the ACS. 

\begin{figure*}[t]
\centering
\subfigure[Entropy of base models' responses, for five of the ACS questions.]{\includegraphics[width=0.85\linewidth]{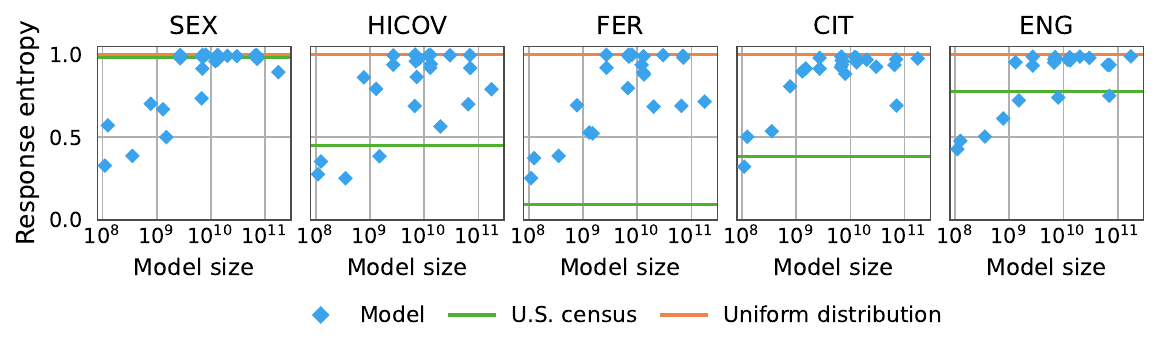}\label{fig:naive1}}
\subfigure[Entropy of base models' responses to the ACS, ordered by model 
size.]{\includegraphics[width=0.9\linewidth]{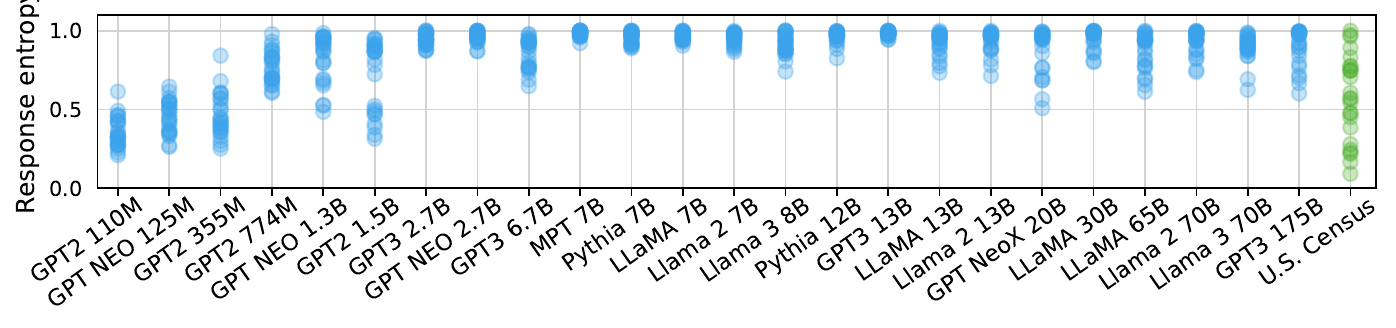}\label{fig:unadjusted}}
\caption{Entropy of model responses across the ACS questions for naive prompting. Entropy of models' responses ({\tiny\textcolor{myblue}{\ding{117}}}) tends to increase log-linearly with model size, irrespective of the underlying response entropy observed in the U.S. census ({\bfseries\textcolor{mygreen}{--}}).}
\end{figure*}

For a first investigation, we consider the normalized entropy of models' responses to the ``SEX", ``HICOV'', and ``FER" questions. The SEX question inquiries about the person's sex, encoded as male female, the HICOV question inquiries whether the person is currently covered by any health insurance plan, and the FER question inquires whether the person has given birth in the past 12 months. When surveying the U.S. population, these three questions elicit responses with very different entropy; responses to the SEX question are almost uniformly distributed, whereas most people answer ``No'' to the FER question.  In contrast, as shown in Figure~\ref{fig:naive1}, the entropy of models' responses to these three questions are surprisingly similar. In particular, we find that the entropy of models' responses tends to increase log-linearly with model size, independent of the question asked. This trend is consistent across all ACS survey questions, see Figure~\ref{fig:var_ent_nonrand} in Appendix~\ref{sec:naiveapp}.

For a broader picture, we illustrate models' response entropy across all survey questions in Figure~\ref{fig:unadjusted}. The blue dots represent models' responses to individual questions, and the green dots represent the entropy of the responses of the U.S. census. We order models by size. We observe that the entropy of responses of the U.S. census greatly varies across questions. In contrast, for any given model, the entropy of its responses varies substantially less so. 

Overall, we find that models' response distributions seem to be widely independent of the survey question asked, and variations across models are much larger than variations across questions. This lead us to suspect that observed differences across models might arise mostly due to systematic biases.~\looseness=-1

\subsection{Testing for systematic biases: A-bias}\label{sec:a-bias}

It is well-known that language models' most likely answer to multiple-choice questions can change depending on seemingly minor factors such as the ordering of few-shot examples~\citep{zhao2021calibrate, lu2022fantastically} or the ordering of answer choices~\citep{robinson2022leveraging}. We are interested in the extent to which changes in choice ordering affect a model's output \emph{distribution over answers}. 

We start by measuring \emph{A-bias}: the tendency of a model towards picking the answer choice labeled ``A". In particular, we seek to study the extent to which the strength of this bias explains the differences in responses observed across models.
For an unbiased model that outputs the same answer distribution irrespective of choice ordering, the expected choice distribution $\bar O^m_q$ under randomized choice ordering would match precisely the uniform distribution (e.g., $\mathrm P$(``A'') = $\mathrm P$(``B'') = 0.5). 
We define a model's A-bias as its absolute deviation from this unbiased baseline:

\begin{equation}
\mathrm{Abias}_q^m := \mathrm P(\bar O_q^{m}=``A") - 1/k_q
\label{eq:abias}
\end{equation}

\begin{figure*}[t]
\centering
\includegraphics[width=\linewidth]{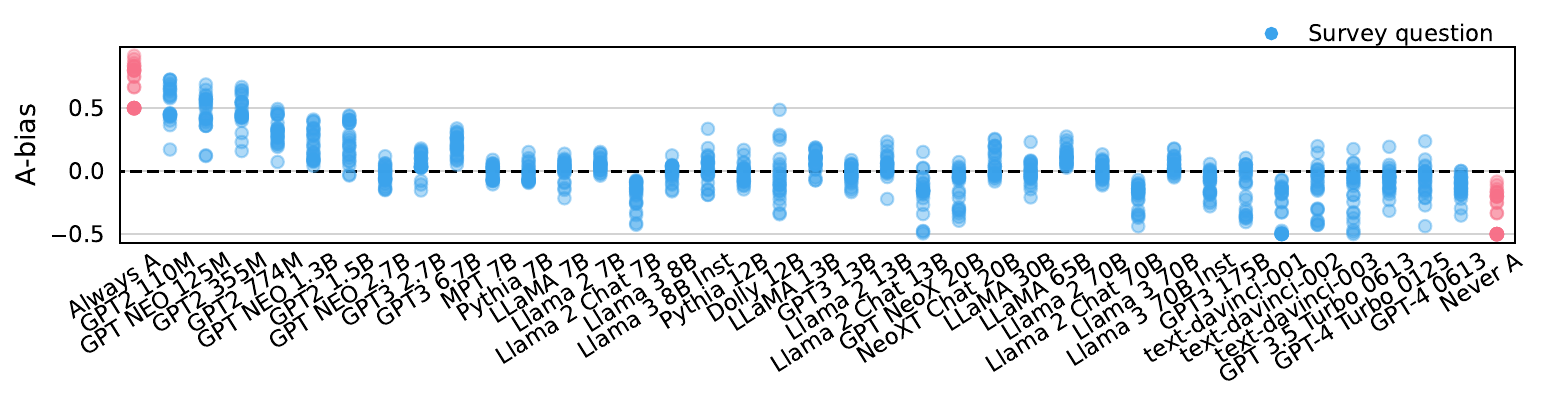}
\caption{A-bias of in model responses across ACS questions. Each dot corresponds to one of the 25 questions. Models are ordered by size. As a reference, the extreme points illustrate A-bias for a model that always answers 'A' and a model that never answers 'A'. All models suffer from substantial A-bias.}
\label{fig:abias}
\end{figure*}

We measure A-bias for each question $q$ and model $m$. Results are illustrated in Figure~\ref{fig:abias}. We again sort models by their size. We observe all models exhibit substantial A-bias. However, models in the order of a few billion parameters or fewer consistently exhibit particularly strong A-bias, and tend towards mono answers. We additionally observe that the strength of A-bias in instruction or RLHF tuned models is similar to that of base models. A plausible explanation for small models exhibiting strong A-bias is that the ability to answer MMLU-style multiple-choice questions only emerges for models of sufficient scale~\citep{dominguez2024training}.

We investigate other types of labeling and position bias (e.g., last-choice bias) in Appendix~\ref{app:randomization}. Overall, we find a strong tendency of LLMs to pick up on spurious signals in the way that answers are ordered and labeled, rather than their semantic meaning. Notably, in contrast to the primacy bias observed in humans
~\citep{groves2009survey}, we find that models exhibit substantial A-bias even when randomizing the position of the ``A'' choice.
Our findings are consistent with the concurrent work of \citet{tjuatja2023llms}, which similarly finds that models' response biases to multiple-choice survey questions are generally not human-like. \citet{wang2024my} additionally shows that models' responses to multiple-choice survey questions may not consistently reflect their free-form outputs.

In summary, we find that systematic biases confound models' answer distributions. 
This makes it challenging to draw robust conclusions about inherent properties of LLMs, such as the opinions or populations they best represent. For example, simply reversing the order of answers to the ``SEX'' question could lead to GPT-2 seemingly representing a population where females are significantly over-represented, whereas a reverse conclusion would be drawn when using the standard answer order. 
While much research went into designing the ACS to elicit faithful answers and eliminate systematic biases when surveying human populations, simply using the same question framing does not protect against the systematic response biases that language models exhibit.

\begin{figure*}[t]
\centering
\subfigure{\includegraphics[width=0.6\linewidth]{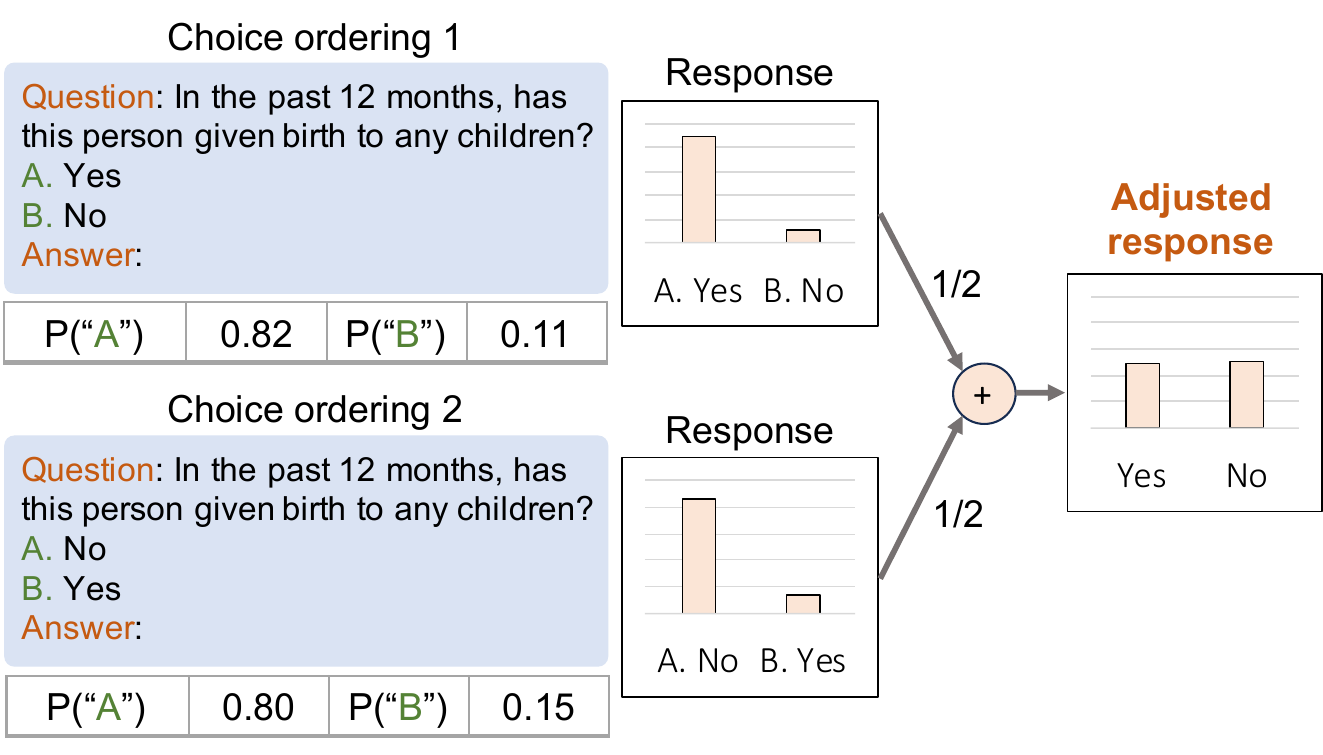}}
\subfigure{\includegraphics[width=
\linewidth]{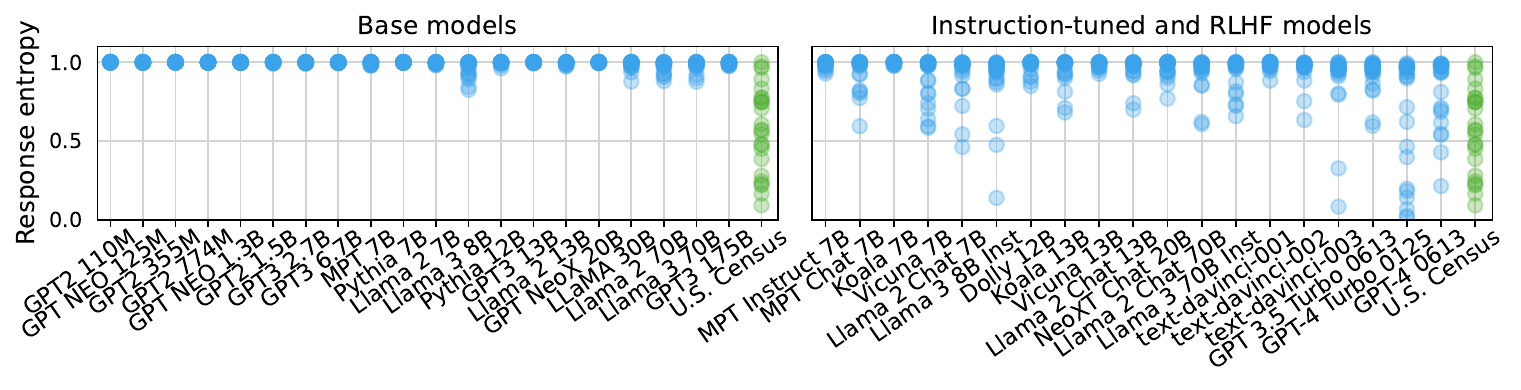}}
\caption{Entropy of model responses after adjustment. \emph{(top)} Illustration of how adjustment is performed. We average models' responses over all possible answer orderings. \emph{(bottom)} Entropy of models' responses after adjustment. Entropy of base models' responses is close to 1 (i.e., uniform). Instruction tuned-models exhibit substantially higher variations in entropy across questions.}
\label{fig:adjusted}
\end{figure*}

\section{Inspecting adjusted responses} 
\label{sec:adjusted}

To eliminate confounding due to labeling and ordering biases, we survey models under randomized choice ordering, borrowing an established methodology to adjust for ordering biases of all kinds in survey research~\citep{groves2009survey}. Also a recent work in LLM research adopts this methodology~\citep{robinson2023leveraging}.
In the following, we refer to the expected response after answer choice randomization as the \emph{adjusted} response.

In Figure~\ref{fig:adjusted} we plot the normalized entropy of models' adjusted responses for the ACS questions considered. First focusing on base models, and comparing the results to Figure~\ref{fig:unadjusted} we find that after adjustment, 1) the variations in responses' entropy across survey questions are very small, 2) we no longer observe the trend of the entropy of model responses increasing log-linearly with model size. In fact, models’ survey responses have a normalized entropy of approximately~$1$ irrespective of model size or survey question asked. This validates our initial hypothesis that, without adjustment, variations in responses across base models arise predominantly due to systematic biases such as A-bias, rather than the content of the survey questions asked.

\subsection{Effect of instruction tuning} 

We now evaluate language models that have been fine-tuned with instructions and/or human preferences, henceforth ``instruction-tuned models''. In the right plot of Figure~\ref{fig:adjusted} we show the normalized entropy of instruction-tuned models' ACS survey responses after adjustment. We observe that instruction tuned-models all exhibit substantially higher variations in entropy across questions compared to base models. But in general, the entropy of their responses remains higher than the entropy of the census responses. Interestingly, as we will see, although deviating more from uniform, model responses do not tend to be closer to the U.S. census responses.

\subsection{Comparing model responses to the U.S. census} 

We now investigate the similarity of language models' adjusted responses to the census data. To do so, we consider the overall U.S. census population, as well as 50 census subgroups corresponding to every state in the United States. This leads to different human reference populations.

\begin{figure*}[t!]
\centering
\includegraphics[width=\linewidth]{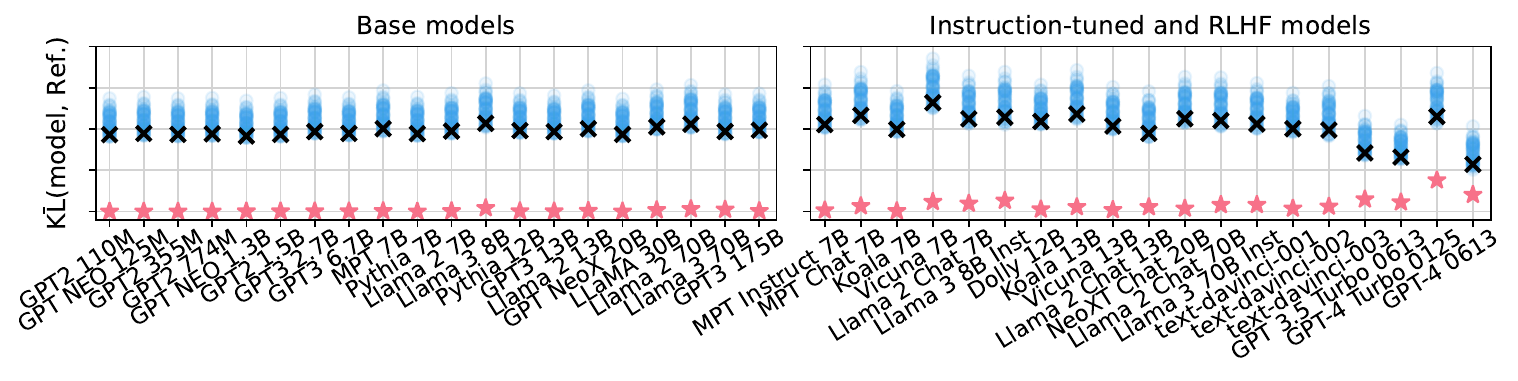}\label{fig:div}
\caption{Divergence between adjusted model responses and different baselines: the overall U.S. census ({\scriptsize\textcolor{black}{\faTimes}}), individual U.S. states ({\tiny\textcolor{myblue}{\ding{108}}}), and a uniform baseline~({\scriptsize\textcolor{myred}{\ding{72}}}). Smaller means more similar.  Model responses are by far  more similar to the uniform baseline than to any human reference population.}
\label{fig:subgroups-mean}
\end{figure*}

Inspired by the alignment measures proposed by \citet{santurkar2023opinions} and \citet{durmus2023towards}, we investigate the similarity of model responses to the census data by evaluating the average divergence across questions between model responses and the census statistics.\footnote{Whereas \citet{santurkar2023opinions} use the Wasserstein distance to compare answer distributions, we use KL divergence since questions in the ACS are predominantly nominal, rather than ordinal.} As we focus on categorical questions, we evaluate average KL divergence between each language model $m$ and each reference population $\mathrm{Ref}$, as follows: 
\[\bar{\mathrm {KL}}(m, \mathrm{Ref})=\frac 1 {|Q|}\sum_{q\in Q} \mathrm{KL}(\bar{R}_q^m||\mathrm{Ref}_q).\]
Results are depicted in Figure~\ref{fig:subgroups-mean}. For each model we plot the divergence to the census in black, the divergence to the different subgroups in blue, and the divergence to a uniform baseline with balanced responses in red. We observe that models are strikingly more similar to the uniform baseline than to any of the populations considered. For base models, this result is unsurprising, since in the previous section we established that base models' responses are essentially uniform after adjustment.

Looking at Figure~\ref{fig:subgroups-mean} we find no consistent trend that instruction-tuning would move responses closer to the census, despite the increased deviation from uniform and the larger variations in entropy (recall Figure~\ref{fig:adjusted}). Only for larger models the divergence seems to clearly decrease with instruction-tuning. However, all models' responses still remain significantly closer to the uniform baseline than to the U.S. census. For instance, for the  GPT-4 model whose answers exhibit the highest similarity to the human reference populations, only 6 out of 25 questions (24\%) are closer to the U.S. census than to the uniform baseline. Given these results, drawing conclusions about the relative alignment of models with subgroups is prone to resulting in brittle conclusions.

\section{Implications for survey-based alignment metrics}
\label{sec:subpopulations}\label{sec:subpops}

\begin{figure*}
\centering
\subfigure[GPT-2 and GPT-Neo.]{\includegraphics[width=0.495\linewidth]{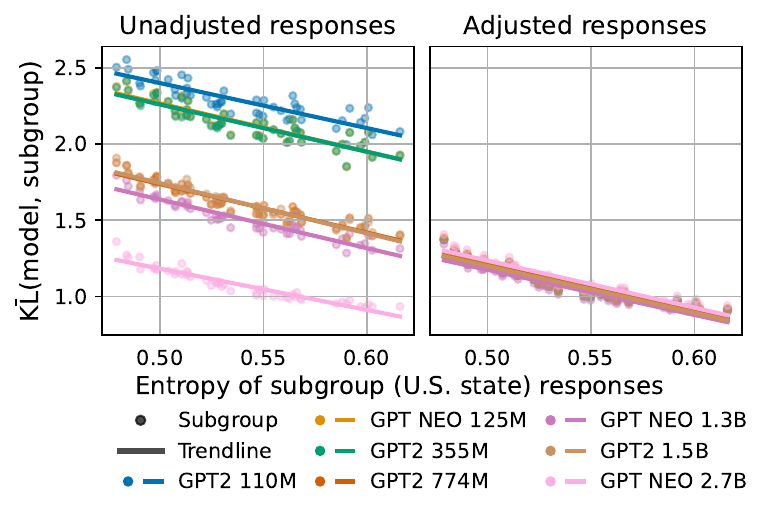}\label{fig:rel-gpt2}}
\hfil
\subfigure[OpenAI's API models.]{\includegraphics[width=0.47\linewidth]{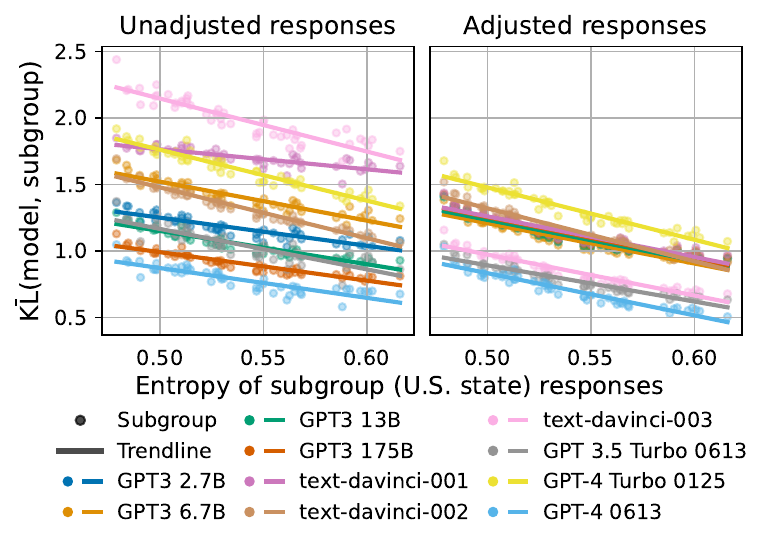}\label{fig:rel-openai}}
\subfigure[MPT, Pythia, GPT-NeoX and its instruction variants.]{\includegraphics[width=0.495\linewidth]{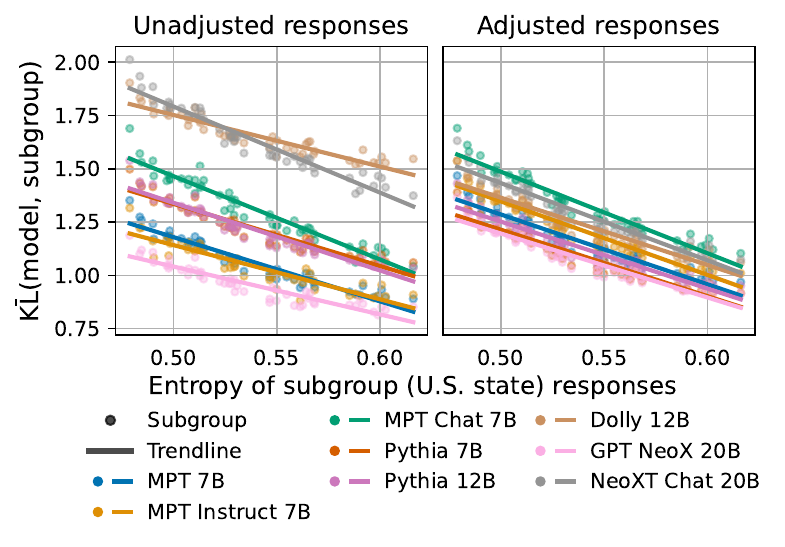}\label{fig:rel-mis}}
\hfil
\subfigure[Llama and its instruction and chat variants.]{\includegraphics[width=0.495\linewidth]{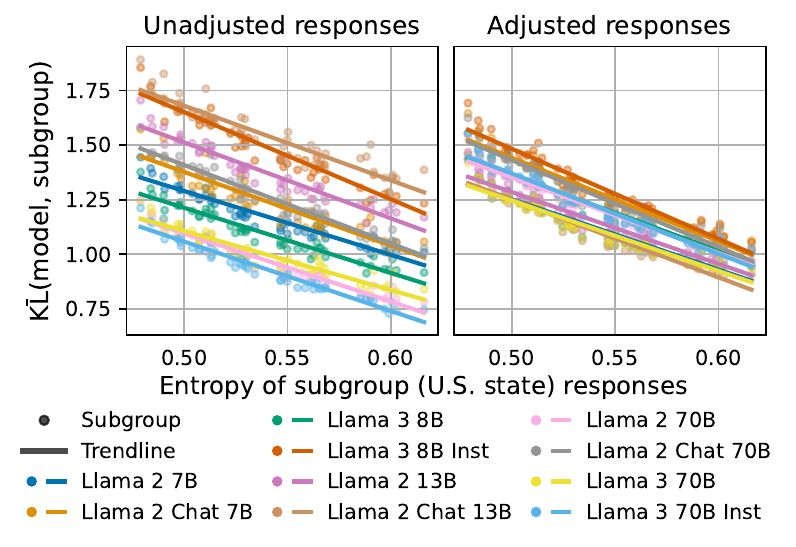}\label{fig:rel-llama}}
\caption{Alignment of models with different census subgroups. All models tend to exhibit similar relative alignment, and the divergence metric decreases with the entropy of the subgroups' responses.}
\label{fig:linAll}
\end{figure*}

Our findings add important context to previous works studying the alignment of language models with different human subpopulations. In particular, we highlighted the tendency of models towards balanced answers. Due to varying entropy in the responses of subgroups this leads to a strong correlation between model alignment and the reference population's entropy. The linear trend in Figure~\ref{fig:linAll} visualizes this. 
For any given model, it consistently {appears} to be more ``aligned'' with the subpopulations exhibiting high entropy in their answers. Interestingly, we find that this trend also holds pre-adjustment, suggesting that the transformation of the response through randomized choice ordering is orthogonal to differentiating aspects of any specific population. 
In contrast, when comparing different models in Figure~\ref{fig:linAll}, we can see how adjustment has a large influence on their relative order. Differences across models that we see under naive prompting disappear after adjustment, which means that are largely attributable to systematic biases, rather than inherent properties of the models.

Taken together our findings imply that the survey-derived alignment measure is more informative of differences in the reference populations rather than the language models is aims to evaluate. Model particularities, such as the pre-training data used, instruction tuning or the use of reinforcement learning with human feedback, seem to have little impact on which population is best represented.~\looseness=-1

\subsection{Beyond the ACS} 

\begin{figure*}
\centering
\subfigure[ATP surveys.]{\includegraphics[width=0.495\linewidth]{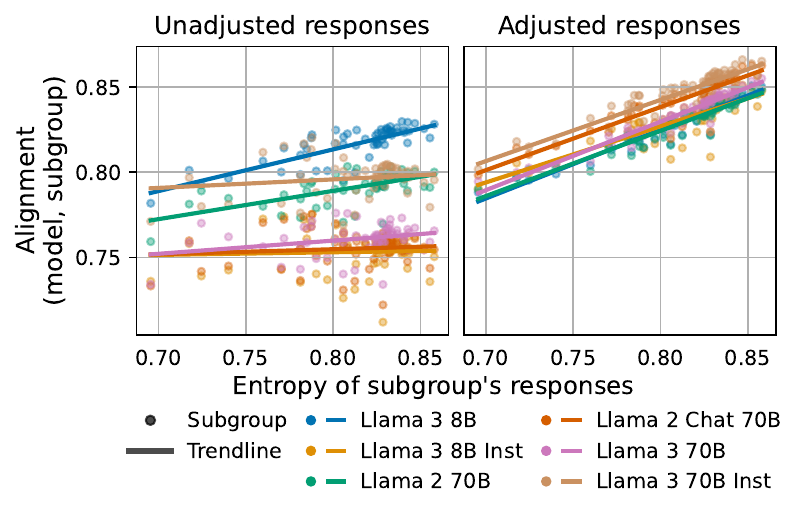}\label{fig:rel-atp}}
\hfil
\subfigure[GAS/WVS surveys.]{\includegraphics[width=0.495\linewidth]{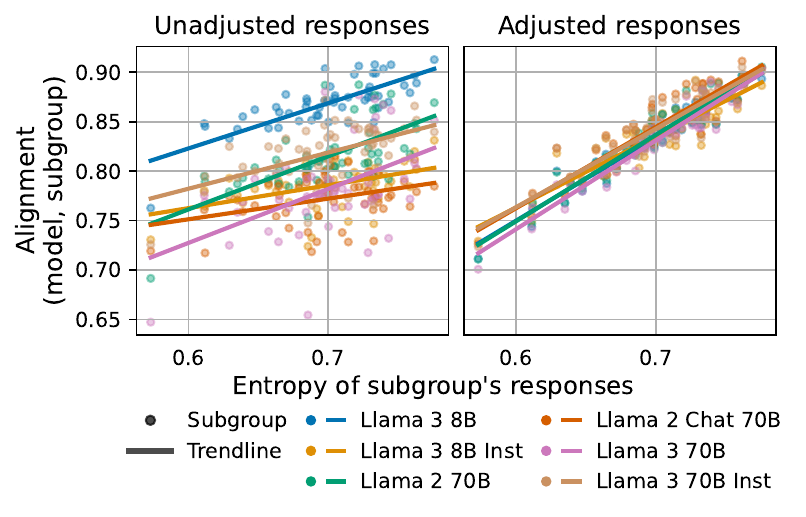}\label{fig:rel-gas}}
\caption{Alignment beyond ACS for selected models. We adopt the measures of \citet{santurkar2023opinions} and \citet{durmus2023towards} on ATP and GAS/VVS opinion surveys. Again, the alignment between models and a given subpopulation correlates with the entropy of the subpopulations' responses.  }
\label{fig:aligment-other}
\end{figure*} 

To inspect whether this trend changes with the content of the questions asked, we reproduce our experiments with  additional surveys. We use the American Trends Panel (ATP) opinion surveys considered by \citet{santurkar2023opinions}, and the Pew Research's Global Attitudes Surveys (GAS) and World Values Surveys (WVS) considered by \citet{durmus2023towards}. These surveys encompass around 1500 questions and 60 U.S. demographic subgroups, and around 2300 questions and 60 national populations, respectively. We adopt the alignment metrics considered by the aforementioned works.
We find that our insights gained from the ACS also hold for the ATP and GAS/WVS surveys.
In particular, we similarly find a linear trend between the alignment metrics and subgroups' entropy of responses, in particular after adjustment, see Figure~\ref{fig:aligment-other}. 

This observation partially explains some of the findings in prior works. For example, \citet{santurkar2023opinions} find that ``all the base models share striking similarities--e.g., being most aligned with lower income, moderate, and Protestant or Roman Catholic groups'' and ``our analysis [...] surfaces groups whose opinions are poorly reflected by current LLMs (e.g., 65+ and widowed individuals)''. For the ATP surveys considered, low income, moderate, and Protestant/Catholic are precisely the demographic subgroups with responses closest to uniformly random among the income, political ideology, and religion demographic subgroups; whereas age 65+ and widowed are the demographic subgroups with responses furthest from uniform among the age and marital status demographic subgroups. 
Further, \citet{santurkar2023opinions} observe that RLHF can result in a ``substantial shift [...] towards more liberal, educated, and wealthy [demographic groups]''. Our results suggest that this could be an artifact of systematic biases. For the ATP surveys, we observe three outliers for which its alignment \emph{before adjustment} is not correlated with the entropy of subgroup's responses: Llama~2 70B Chat and the two Llama~3 Instruct models. These are the models with largest pre-training compute considered. However, after adjustment, the alignment trends of Llama 2 70B Chat and the Llama 3 Instruct models are remarkably similar to that of their corresponding base models and all other LLMs.

\section{Conclusion}
\label{sec:discussion}

We used a popular methodology to elicit LLMs' answer distributions to survey questions and closely examined the  responses on the basis of the prime US demographic survey.  We found that model responses are dominated by systematic ordering biases and do not exhibit the natural variations in entropy found in the human reference data collected by the US census. Even after adjusting for ordering biases, LLMs' responses still do not resemble those of human populations. Instead, they exhibit consistently high entropy, independent of the question asked. This holds true irrespective of model size or fine-tuning with human preferences.~\looseness-1

These findings have important implications for insights gained from survey-derived alignment metrics. In particular, it explains why models of varying size all exhibit the same trend: they are most aligned with subgroups who happen to have balanced answers for the survey questions under consideration. For all models and surveys considered, alignment appears to be a proxy for the entropy of subgroups, rather than an inherent property of the model, or its training data. 

We want to reiterate that our focus lies on questioning a popular methodology of eliciting survey responses from large language models using multiple choice prompting. At the example of this methodology our results highlight an important pitfall and suggest caution to expect robust insights when comparing such responses against those of human populations. The robustness and quality of an established survey does not seamlessly translate from the results obtained by surveying human populations to the logits output by LLMs. More research is needed to design methodologies for getting insights into the inherent biases of LLMs and the population they might represent. Here public surveys and their accompanying data offer exciting potential and the could play an important role as a benchmarking tool for systematic evaluations of LLMs, see \citep{cruz2024evaluating} as an example. Although the use of survey data for LLM research has recently gained  popularity, it still remains a widely under explored data source.~\looseness=-1

\section*{Acknowledgements}

The authors would like to thank Frauke Kreuter and the Social Data Science and AI Lab at Ludwig-Maximilians-Universität Munich for inspiring discussions on an earlier version of this manuscript. 
Celestine Mendler-Dünner acknowledges financial support from the Hector Foundation. 

\bibliographystyle{apalike}
\bibliography{references}

\newpage
\appendix
\onecolumn

\section{Experimental details}

We use the American Community Survey (ACS) Public Use Microdata Sample (PUMS) files made available by the U.S. Census Bureau.\footnote{\href{https://www.census.gov/programs-surveys/acs/microdata.html}{https://www.census.gov/programs-surveys/acs/microdata.html}} The data itself is governed by the terms of use provided by the Census Bureau.\footnote{\href{https://www.census.gov/data/developers/about/terms-of-service.html}{https://www.census.gov/data/developers/about/terms-of-service.html}} We download the data directly from the  U.S. Census using the Folktables Python package~\citep{ding2021retiring}. We download the files corresponding to the year 2019.

We downloaded the publicly available language model weights from their respective official HuggingFace repositories. We run the models in an internal cluster. The total number of GPU hours needed to complete all experiments is approximately 1500 (NVIDIA A100). The budget spent querying the OpenAI models was approximately \$200.

We open source the code to replicate all experiments.\footnote{\href{https://github.com/socialfoundations/surveying-language-models}{https://github.com/socialfoundations/surveying-language-models}} 
In addition, the repository contains notebooks to visualize the results of our investigations under different prompt ablations. 

\subsection{Survey questionnaire used}\label{sec:questionnaire}

The exact questionnaire used in our experiments can be retrieved from our Github repository. 
We consider 25 questions from the 2019 ACS questionnaire corresponding to the following variables in the Public Use Microdata Sample: SEX, AGEP, HISP, RAC1P, NATIVITY, CIT, SCH, SCHL, LANX, ENG, HICOV, DEAR, DEYE, MAR, FER, GCL, MIL, WRK, ESR, JWTRNS, WKL, WKWN, WKHP, COW, PINCP. We take all questions as they appear in the ACS, with the exceptions:
\begin{itemize}
    \item HISP: The ACS contains 5 answer choices corresponding to different Hispanic, Latino, and Spanish origins, and respondents are instructed to write down their origin if their origin is not among the choices provided. We instead provide two choices: ``Yes'' and ``No".
    \item RAC1P: The ACS contains 15 answer choices, allows for selecting multiple choices, and respondents are instructed to write down their race if not among those in the multiple choice. The PUMS then provides up to 170 race codes (RAC2P and RAC3P). We instead present 9 choices, corresponding to the race codes of the RAC1P varible in the PUMS data dictionary.
\end{itemize}

Additionally, the variables ESR and COW are not directly associated with any single question in the ACS, but rather aggregate employment information. We formulate them as questions by taking the PUMS data dictionary's variable and codes descriptions. Lastly, for the questions corresponding to the variables AGE, WKWN, WKHP, and PINCP, respondents are asked to write down an integer number. We convert such questions to multiple-choice via binning. 

\section{Detailed experimental results}

The results in this section complement Section~3, and pertain non-instruction-tuned language models. 

\subsection{Model responses across questions before adjusting for A-bias}

\label{sec:naiveapp}

\begin{figure}
    \centering
    \includegraphics[width=0.95\linewidth]{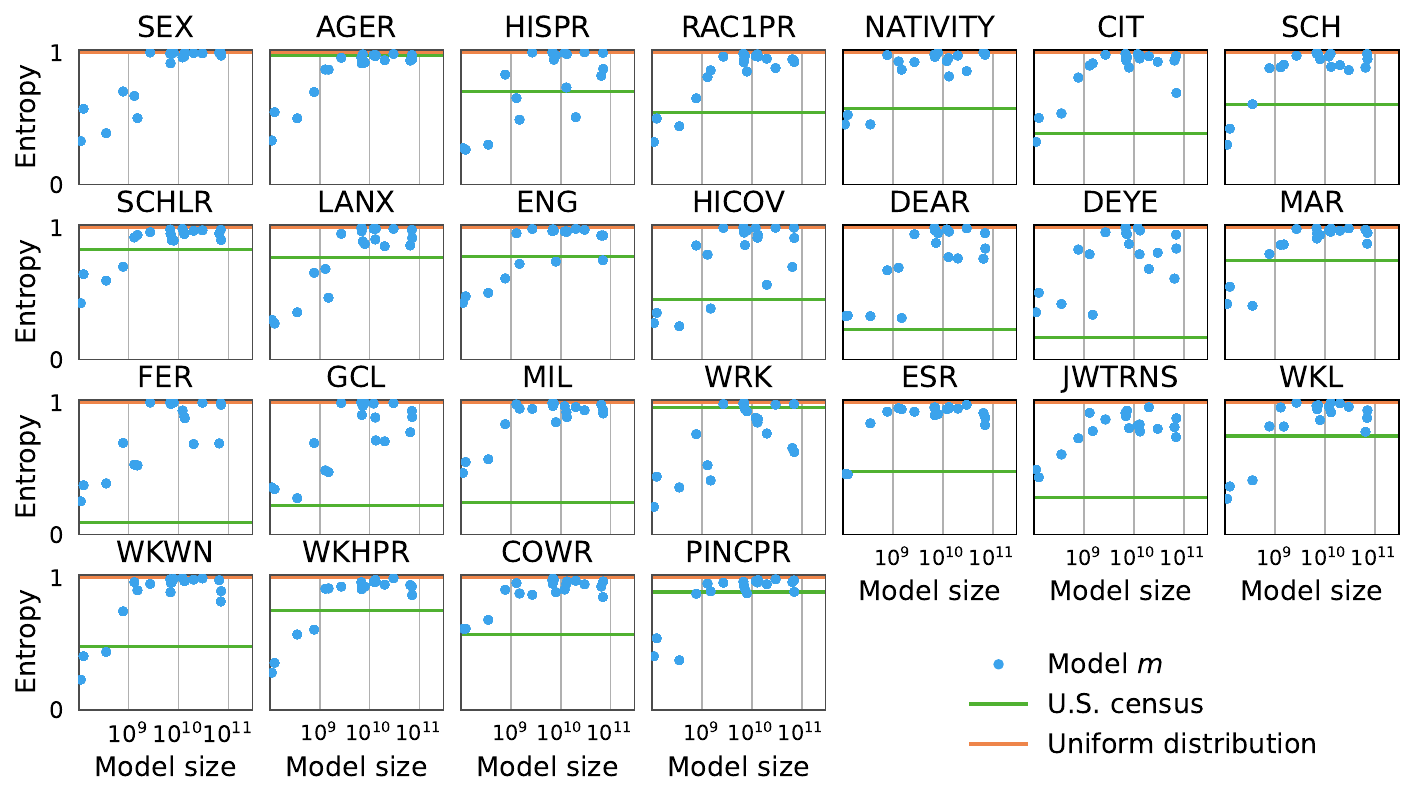}
    \caption{Normalized entropy of survey responses for individual questions (without adjustment).}
    \label{fig:var_ent_nonrand}
\end{figure}

When surveying models without choice order randomization, we observe that the entropy of model responses tends to increase log-linearly with model size, often matching the entropy of the uniform distribution for the larger models. This trend is consistent across survey questions, irrespective of the question's distribution over responses observed in the U.S. census, see Figure~\ref{fig:var_ent_nonrand}. 

\section{Ordering bias: further experiments}\label{app:randomization}

We conduct additional randomization experiments pertaining to answer choice position and labeling bias, complimenting Section~3. We consider the GPT-2, GPT Neo, MPT, Pythia, and LLaMA models. The experiments follow a consistent setup:

\begin{enumerate}[topsep=0pt,itemsep=0pt]
    \item We randomize both the order in which choices are presented and the label (i.e., letter) assigned to each answer choice. For example, for the "sex" question, the possible combinations are ``A. Male B. Female'', ``A. Female B. Male'', ``B. Male A. Female'', and ``B. Female A. Male''. Note that in the experiments presented in Section~\ref{sec:a-bias} we only randomized over the order in which choices are presented (i.e., the ``A'' choice was always presented first).
    \item We compute the output distribution over responses for choice position (the probability assigned to the first, second, etc., answer choice presented) and letter assignment (the probability assigned to the answer choice assigned ``A'', ``B'', etc.).
\end{enumerate}

For each model and survey question, we estimate the expected distribution over responses for both choice position and letter assignment by collecting 3,000 responses (step 2) under different randomizations of choice position and letter assignment (step 1). A model with no position and labeling biases would assign the same probability distribution to answer choices (e.g., ``male'' and ``female'') regardless of position or letter assignment, and therefore the expected distributions over position (e.g., selecting the first choice) and letter assignment (e.g., selecting ``A'') would be uniform. 

\subsection{Disentangling ordering bias into positioning bias and labeling bias}

We perform chi-square tests to determine whether language models' output responses distributions over position and letter assignment significantly deviate from the uniform distribution (i.e., if there exists statistically significant bias in position or letter assignment). Since we collect 3,000 response distributions under randomized choice position and letter assignment, we ensure a high test power ($\geq$ 0.98) in detecting small effect sizes (0.1) at a significance level of 0.05.

\begin{figure}[h!]
  \centering
  \subfigure[Significant ordering bias.]{
    \centering
    \includegraphics[width=0.25\linewidth]{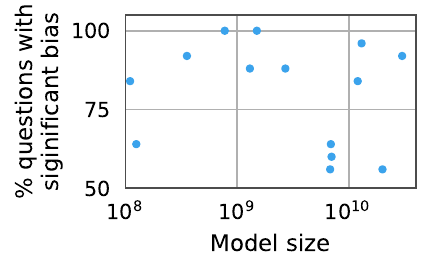}}
    \hfil
  \subfigure[Significant label bias.]{
    \centering
    \includegraphics[width=0.25\linewidth]{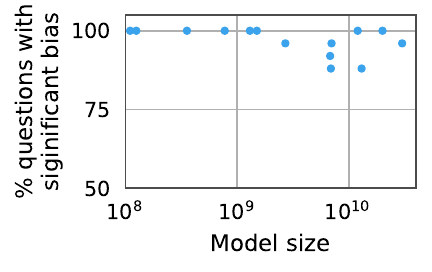}
    }
  \caption{All models exhibit statistically significant letter and ordering bias for most survey questions.}
  \label{fig:abc-bias}
\end{figure}

\begin{figure}[h!]
  \centering
  \subfigure[A-bias]{
    \includegraphics[width=0.47\textwidth]{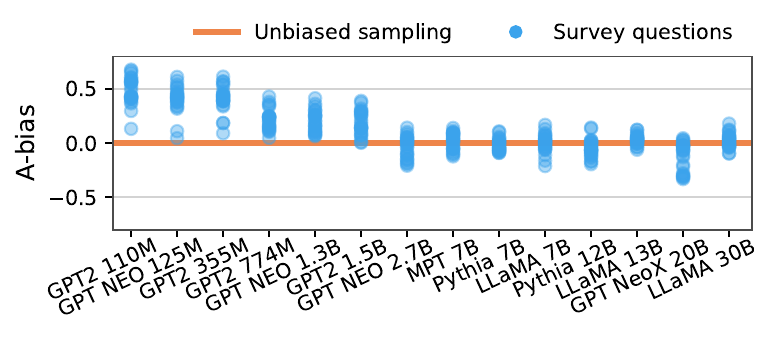}
    }
  \subfigure[First choice bias]{
    \includegraphics[width=0.47\textwidth]{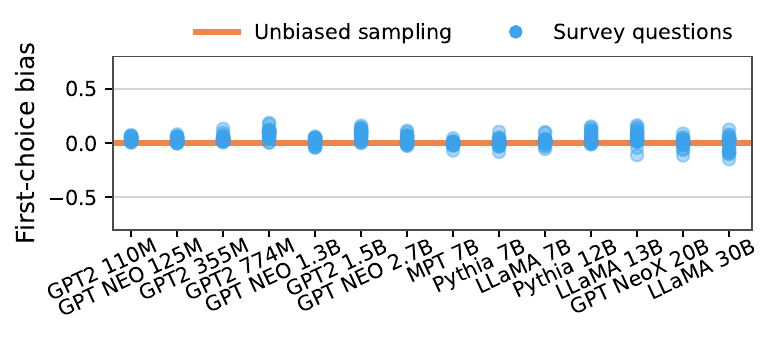}
    }
  \caption{Models, particularly those with less than a few billion parameters, tend to exhibit stronger A-bias than first-choice bias.}
\label{fig:order-bias}
\end{figure}

We find that models exhibit significant positioning and labelling for most survey questions, see Figure~\ref{fig:abc-bias}. We observe that labelling is more prevalent that positioning bias. While both tend to decrease with model size, order bias decreases more significantly with model size, whereas labeling bias tends to be very prevalent across all model sizes. In Figure~\ref{fig:order-bias} we plot both the strength of A-bias and first-choice bias across survey questions. The strength of A-bias tends to be greater than that of first-choice bias, particularly for the smaller models.

\subsection{I-bias} 
\begin{figure}
  \centering
  \subfigure[A-bias in the ``A'', ``I'' randomization experiment.]{
    \includegraphics[width=0.47\linewidth]{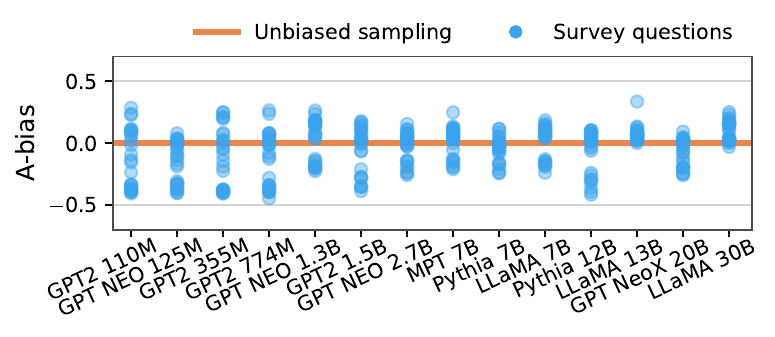}
    }
    \subfigure[I-bias in the ``A'', ``I'' randomization experiment.]{
    \includegraphics[width=0.47\linewidth]{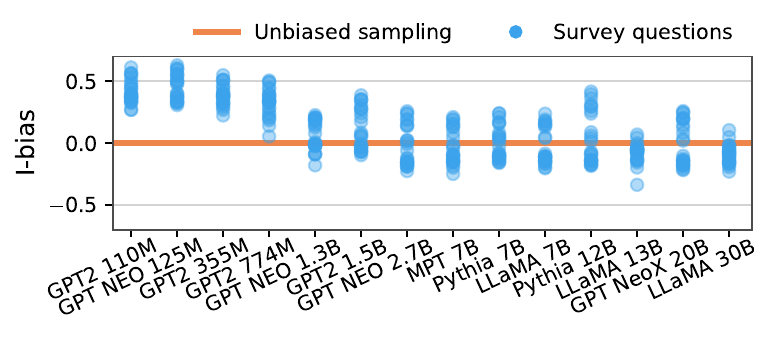}}
  \caption{When both ``A'' and ``I'' are present, small models exhibit I-bias rather than A-bias.}
\label{fig:ai-bias}
\end{figure}

We hypothesize that A-bias is prevalent because the single character ``A'' is relatively frequent as the starting word of a sentence in written English. We test this hypothesis by replacing the character ``B'' with ``I'' when presenting the survey questions, since the character ``I'' is even more frequent as the starting word of a sentence in written English. We randomize over choice ordering and label assignment as in the previous evaluation. We find that, when presenting both ``A'' and ``I'', small models then exhibit I-bias rather than A-bias (Figure~\ref{fig:ai-bias}), supporting our initial hypothesis.

\subsection{Using letters with similar frequency in written English} 
\begin{figure}
  \centering
  \subfigure[Significant labelling bias.]{
  \centering
    \includegraphics[width=0.22\linewidth]{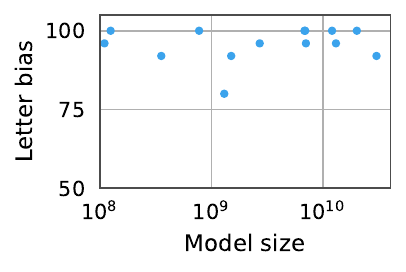}}
    \hfil
  \subfigure[Significant ordering bias.]{
  \centering
    \includegraphics[width=0.22\linewidth]{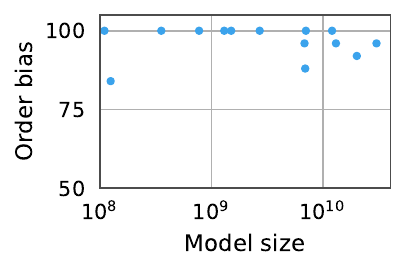}
  }
  \caption{``R'', ``S'', ``N'', etc. randomization experiment. All models, irrespective of size, exhibit statistically significant letter and positioning bias for most survey questions.}
\label{fig:rsn-bias}
\end{figure}

Motivated by the I-bias experiment, we now examine whether labeling bias can be mitigated by using letters that have similar frequency in written English. Therefore, instead of assigning to choices the labels ``A'', ``B'', etc. we assign the following labels: ``R'', ``S'', ``N'', ``L'', ``O'', ``T'', ``M'', ``P'', ``W'', ``U'', ``Y'', ``V''. We find that, compared to the ``A'', ``B'', etc. randomization experiment, the percentage of questions for which models exhibit significant labeling bias somewhat decreases (Figure~\ref{fig:rsn-bias}). However, models tend to exhibit substantially more position bias. This indicates that, in the absence of a label that provides a strong signal (e.g., ``A'' or ``I''), models tend to exhibit significantly higher choice-ordering bias, irrespective of model size.

\section{Prompt ablations}\label{sec:ablation-prompt}

We reproduce our experiments using different prompts to query the model. Due to the cost of querying OpenAI's models, we only perform these ablations for models with publicly available weights. The notebooks with all figures can be retrieved from
our Github repository.
\footnote{\href{https://github.com/socialfoundations/surveying-language-models/blob/main/prompt-ablations}{https://github.com/socialfoundations/surveying-language-models/blob/main/prompt-ablations}}

Overall, the prompt ablation results are very consistent with the findings presented in the main text of the paper. In the following we provide an overview over the different ablations performed.

\subsection{System rompt used for GPT-3.5 and GPT-4}

When querying GPT-3.5, GPT-4, and GPT-4 Turbo, we use the system prompt \texttt{\small Please respond with a single letter.}, as otherwise for most questions none of the top-5 logits correspond to answer choice labels (e.g., ``A'', ``B''). Note that this problematic arises due to the fact that the OpenAI API only allows access to the top 5 logits. We adapt the system prompt used by \citet{dorner2023personality} in the context of surveying GPT-4 with standarized personality tests.

\subsection{Individual survey questions}
First, we use different styles to prompt individual survey questions. We enumerate the prompt styles as (P1)-(P8). 

\paragraph{Additional context.} We first explore whether including additional context signaling that the questions presented are from the American Community Survey, or that they are to be answered by U.S. households. Keeping identical survey questions, we append at the start of the prompt one of the following sentences:
\begin{enumerate}[itemsep = 0ex]
    \item[(P1)] \texttt{\small Bellow is a question from the American Community Survey.}
    \item[(P2)] \texttt{\small Answer the following question from the American Community Survey.}
    \item[(P3)] \texttt{\small Answer the following question as if you lived at a household in the \\United States.}
\end{enumerate}

\paragraph{Asking questions in the second person.} We change the framing of the questions.
\begin{enumerate}[itemsep = 0ex]
    \item[(P4)] We modify the survey questionnaire such that questions are formulated in the second person rather than the third person (e.g., ``What is your sex?'' instead of ``What is this person's sex?''). 
\end{enumerate}

\paragraph{Including instructions.} Following the prompt ablation of \citet{santurkar2023opinions}, we append at the start of the prompt one of the following instructions:

\begin{enumerate}[itemsep = 0ex]
    \item[(P5)] \texttt{\small Please read the following multiple-choice question carefully and \\select ONE of the listed options.}
    \item[(P6)] \texttt{\small Please read the multiple-choice question below carefully and \\ select ONE of the listed options. Here is an example of the \\ format:\textbackslash nQuestion: Question 1\textbackslash nA. Option 1\textbackslash nB. Option 2\textbackslash n\\C. Option 3\textbackslash nAnswer: C}
\end{enumerate}

\paragraph{Chat-style prompt.} We consider the prompt used by \citet{durmus2023towards}:
\begin{enumerate}
    \item[(P7)]\texttt{Human: \{question\}\textbackslash nHere are the options:\textbackslash n\{options\}\textbackslash n\\Assistant: If had to select one of the options, my answer \\would be}
\end{enumerate}

\paragraph{Interview-style prompt.} We consider the prompt used by \citet{argyle2022out}:
\begin{enumerate}
    \item[(P8)]\texttt{Interviewer: \{question\}\textbackslash n\{options\}\textbackslash nMe:}
\end{enumerate}

\subsection{Sequential generation}\label{sec:ablation-seq}

We use different prompts to integrate a model's previous responses when prompting subsequent survey questions. Instead of summarizing previous responses using bullet points as in Section~5, we keep previous questions and answers in-context.

\paragraph{Question answering.} Keeping questions and answers in-context resembles the typical few-shot Q\&A setting. For instance, prompting for the third question in the questionnaire corresponds to

\texttt{Question: \{question 1\}\textbackslash n\{options 1\}\textbackslash nAnswer:\{answer 1\}\textbackslash n}

\texttt{Question: \{question 2\}\textbackslash n\{options 2\}\textbackslash nAnswer:\{answer 2\}\textbackslash n}

\texttt{Question: \{question 3\}\textbackslash n\{options 3\}\textbackslash nAnswer:}

\paragraph{Interview-style prompt.} We consider the prompting style used by \citet{argyle2022out}. For instance, prompting for the third question in the questionnaire corresponds to

\texttt{Interviewer: \{question 1\}\textbackslash n\{options 1\}\textbackslash nMe:\{answer 1\}\textbackslash n}

\texttt{Interviewer: \{question 2\}\textbackslash n\{options 2\}\textbackslash nMe:\{answer 2\}\textbackslash n}

\texttt{Interviewer: \{question 3\}\textbackslash n\{options 3\}\textbackslash nMe:}

\section{Results for ATP, GAS, WVS, and ANES surveys}\label{sec:ablation-atp}\label{sec:anes}

We reproduce the experiments of Sections~3 and 4 using the ATP, and GAS/WVS used by \citet{santurkar2023opinions} and \citet{durmus2023towards}, where questions are presented individually of one another. We additionally reproduce the experiments of Section~5 using the 2016 ANES questionnaire considered by \citet{argyle2022out}, where questions are presented in sequence. We do not consider OpenAI's models as the cost to reproduce the experiments via the OpenAI API exceeds our budget. We obtain very similar results to those of the ACS presented in the main text of the paper. The notebooks with all figures can be retrieved from our Github repository.
\footnote{\href{https://github.com/socialfoundations/surveying-language-models}{https://github.com/socialfoundations/surveying-language-models}}

\subsection{ATP surveys}

\begin{figure}
    \centering
    \includegraphics[width=0.50\linewidth]{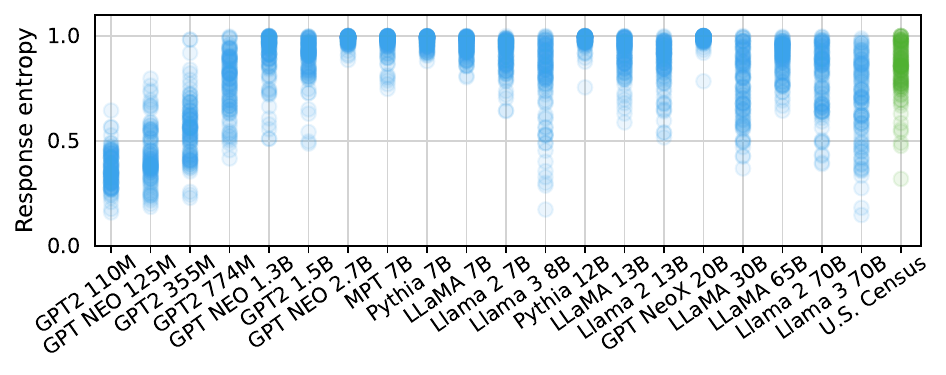}
    \includegraphics[width=0.48\linewidth]{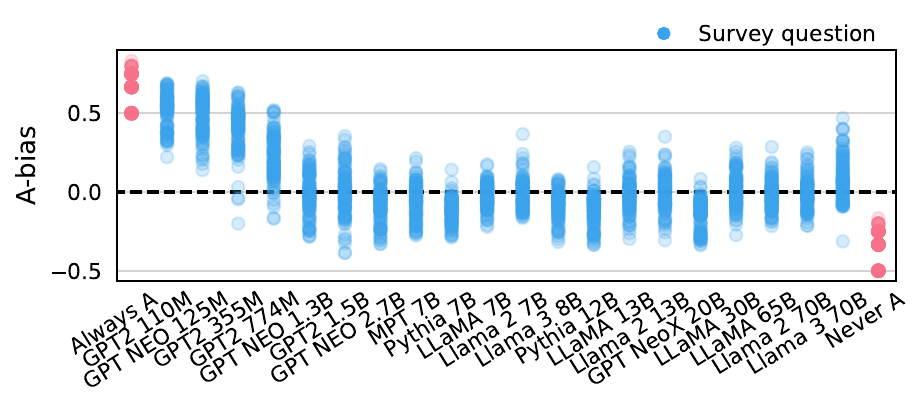}
    \includegraphics[width=0.50\linewidth]{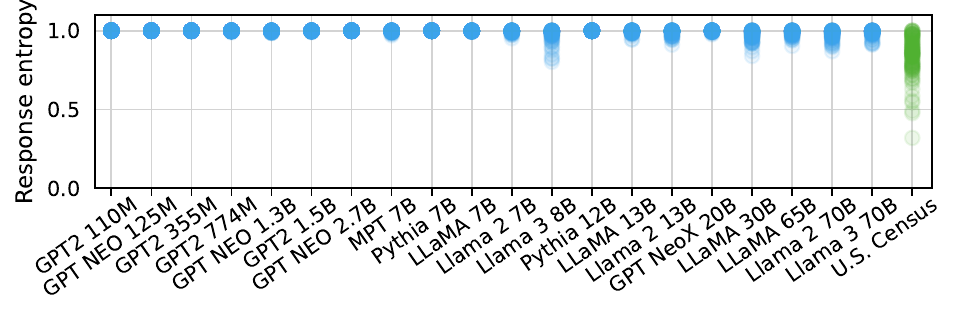}
\includegraphics[width=0.48\linewidth]{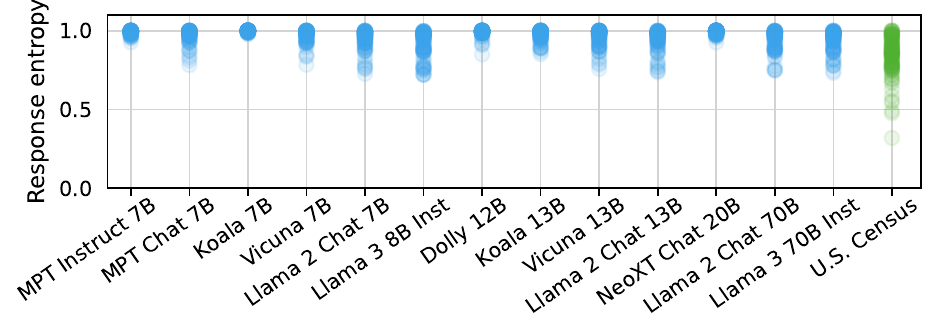}
\includegraphics[width=0.50\linewidth]{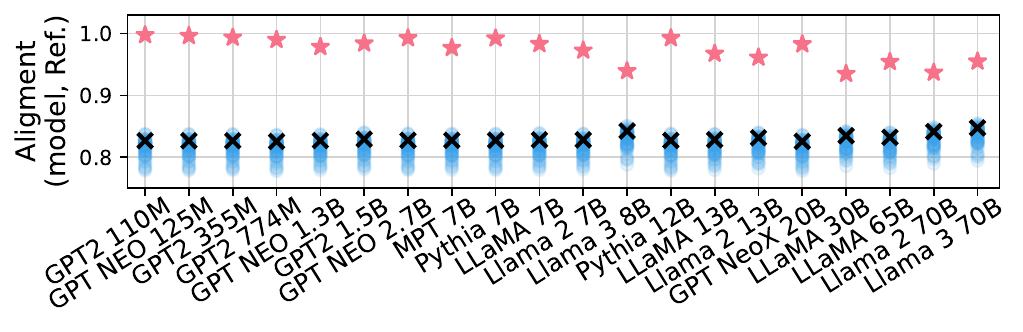}
\includegraphics[width=0.48\linewidth]{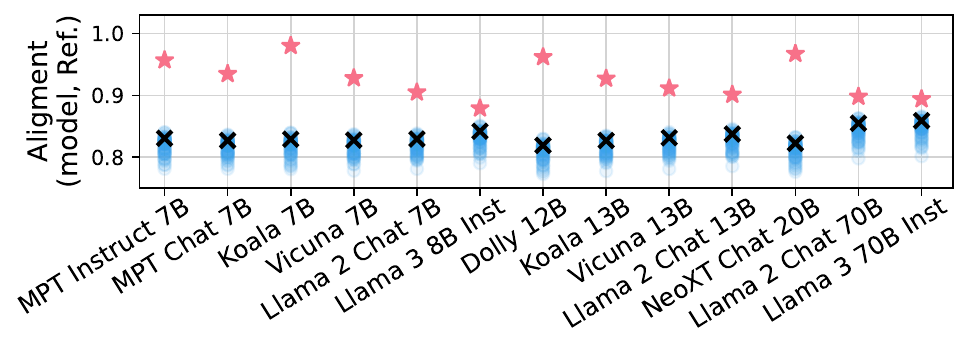}
\caption{Reproduction of the experiments in Sections~3 and 4 for the ATP surveys.}
\label{fig:atp-all}
\end{figure}

We obtain the ATP survey questions and their corresponding human responses from the OpinionsQA repository.\footnote{\url{https://github.com/tatsu-lab/opinions_qa}} We present all answer choices when querying the models, but exclude the answer choices corresponding to refusals from our analysis similarly to \citet{santurkar2023opinions}. When comparing the similarity of models' responses to different demographic subgroups, we use the demographic subgroups and the alignment metric considered by \citet{santurkar2023opinions}. For such metric, higher values of alignment indicate that models' responses are more similar to the reference demographic group. We find that all models are more ``aligned'' with the uniformly random baseline than with any of the demographic subgroups, see Figure~\ref{fig:atp-all}.

\subsection{GAS and WVS surveys}

\begin{figure}
    \centering
        \includegraphics[width=0.49\linewidth]{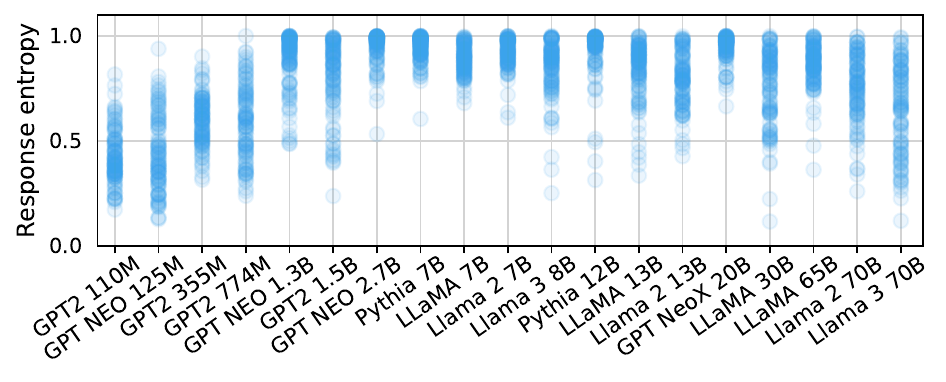}
    \includegraphics[width=0.49\linewidth]{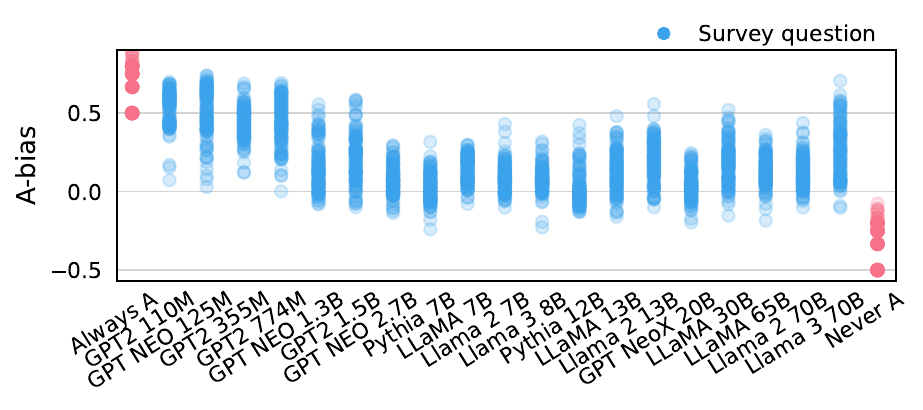}
    \includegraphics[width=0.49\linewidth]{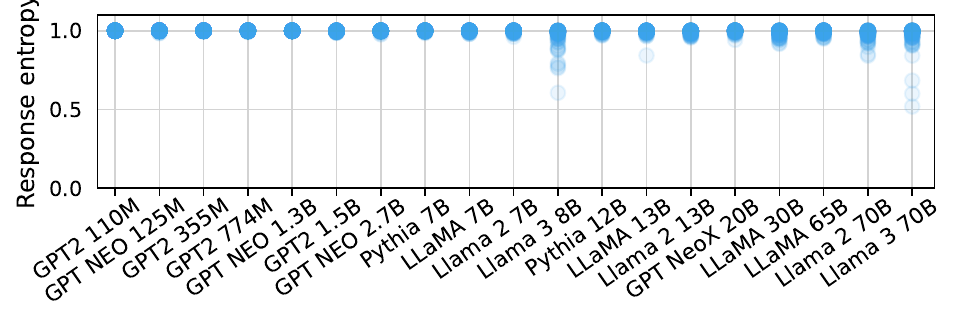}
\includegraphics[width=0.49\linewidth]{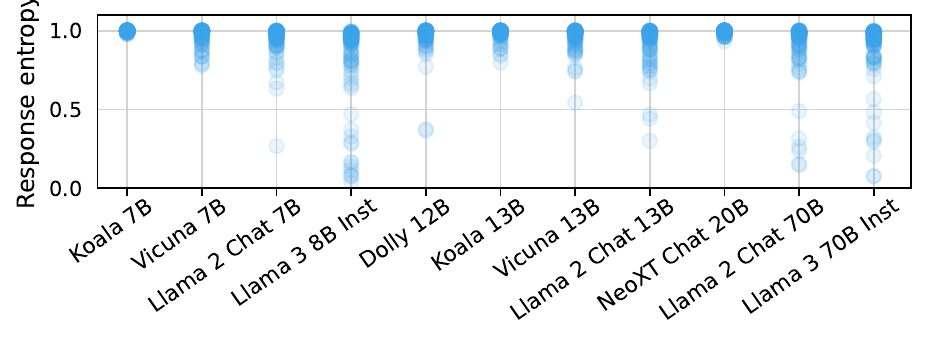}
\includegraphics[width=0.49\linewidth]{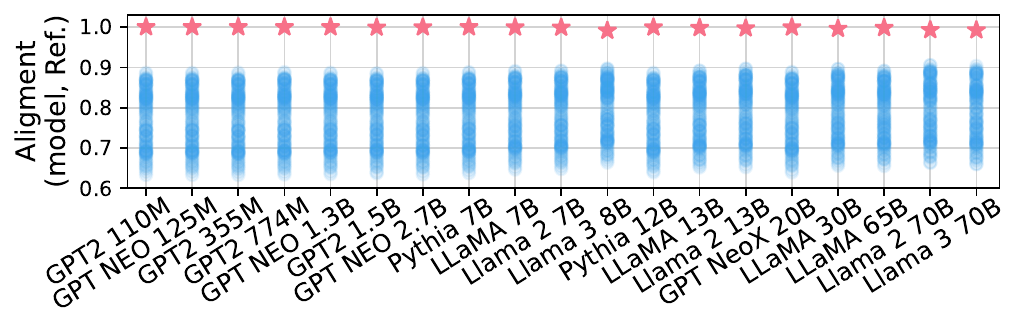}
\includegraphics[width=0.49\linewidth]{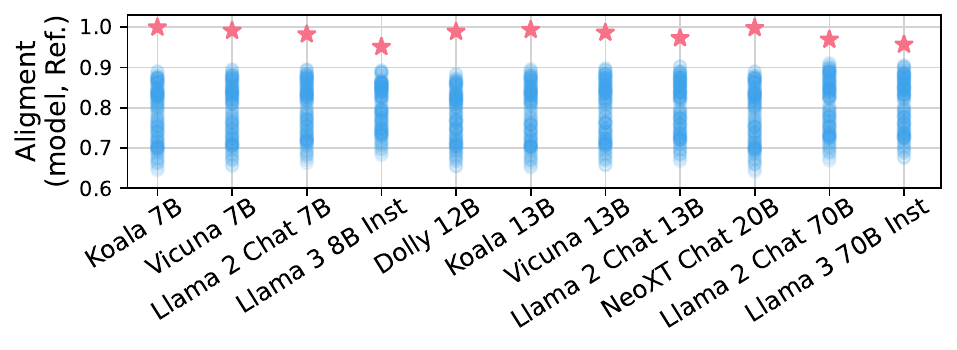}
\caption{Reproduction of the experiments in Sections~3 and 4 for the GAS/WVS surveys.}
\label{fig:gas-all}
\end{figure}

We obtain the ATP survey questions and their corresponding human responses from the GlobalOpinionsQA repository.\footnote{\url{https://huggingface.co/datasets/Anthropic/llm_global_opinions}}
When comparing the similarity of models' responses to the population-level survey responses of different countries, we use the countries and the similarity metric considered by \citet{durmus2023towards}. We find that all models produce survey responses that are more similar to those of the uniformly random baseline than to those of any of the demographic subgroups, see Figure~\ref{fig:gas-all}.

\subsection{Relative alignment for ATP and GAS/WVS surveys}\label{sec:app-rel-atp}

We consider the alignment measures proposed by \citet{santurkar2023opinions} and \citet{durmus2023towards} on ATP and GAS/VVS opinion surveys for the largest base / instruct models considered. We find that, similarly to our observations for the ACS, the alignment between models and  a given subpopulation is highly correlated with the entropy of the subpopulations' responses.

\subsection{ANES survey}

\begin{figure}
\centering
\includegraphics[width=1\linewidth]{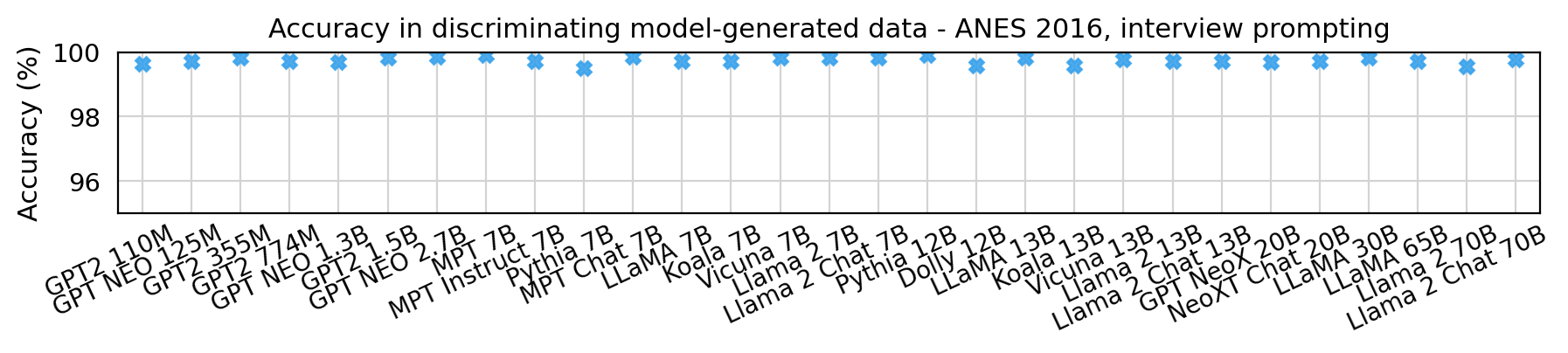}
\caption{The discriminator test performed on datasets generated using the 2016 ANES survey questionnaire (with choice randomization).}
\label{fig:prediction-arg}
\end{figure}

We present questions in the multiple-choice format described in Section~2, using the \texttt{\small Interviewer:}, \texttt{\small Me:} prompt style described by \citet{argyle2022out}.
We retrieve the 2016 ANES data from the official website\footnote{\url{https://electionstudies.org/data-center/2016-time-series-study/}}, and process it such that it matches in form the questionnaire designed by \citet{argyle2022out}. We find that the trained classifiers can discriminate between the model-generated data and the ANES data with very high accuracy ($\geq$99\%), see Figure~\ref{fig:prediction-arg}.

\section{Sequential sampling of responses}\label{sec:generation}

\begin{figure}[t!]
\centering
\includegraphics[width=0.98\linewidth]{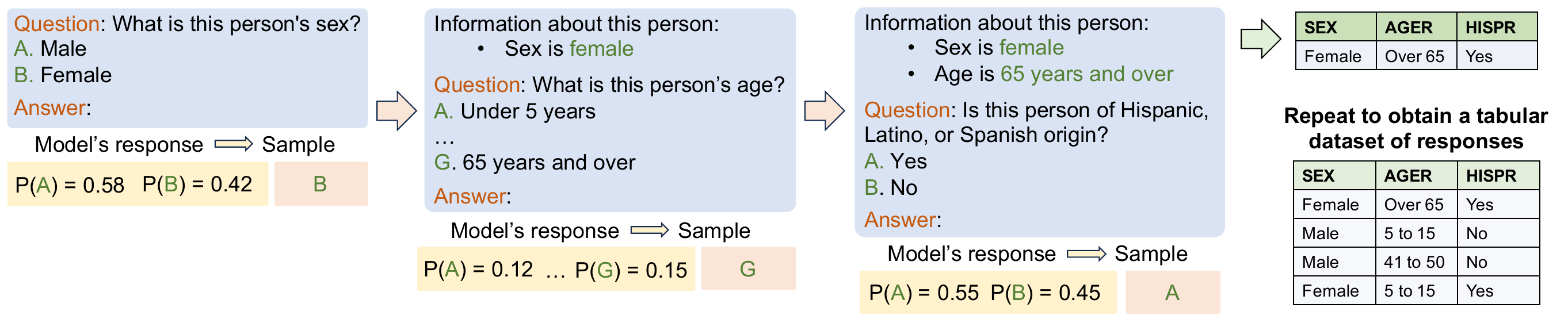}
\caption{Methodology and prompt template used to sequentially sample models' responses to entire survey questionnaires. We provide the answers to previous question in context when prompting subsequent questions. The output is a tabular dataset of responses.}
\label{fig:prompt}
\end{figure}

Motivated by recent findings of~\citet{argyle2022out} we conducted an additional investigation where we seek to fill entire ACS questionnaires in a sequential manner, in order to generate for each language model a synthetic dataset of responses. This data  emulates in form the ACS dataset collected by the U.S. Census Bureau. We then study the extent to which such synthetic datasets resemble the ACS dataset.

\subsection{Methodology} 
We present survey questions in the same order as in the ACS questionnaire. When querying a model to answer survey question $q$, we include a summary of the $q-1$ previously sampled answers in context. \footnote{The maximum number of tokens in a filled questionnaire was less than 1024 tokens in all cases, thus fitting entirely within the context window of all surveyed models.}
We then sample from the model's output probability distribution over answers, and continue to the next question. We illustrate this sequential process in Figure~\ref{fig:prompt}. We refer to Appendix~\ref{sec:ablation-seq} for results collected with different variations of how a model's previous answers are integrated into the prompt. We find our results to be robust to these prompt variations.

For each language model we sample $N$=100,000 model-generated responses to the ACS. Due to the cost of querying OpenAI's models, we only survey GPT-4 and sample $N=500$ responses.
As a result, we generate for each language model a tabular dataset similar in form to the ACS data, with $N$ rows corresponding to each filled questionnaire and 25 columns corresponding to each question. 

\subsection{The discriminator test}\label{sec:disc}

We investigate whether the model-generated datasets resemble the U.S. census data by constructing a binary prediction task aiming to discriminate synthetic responses from census responses. Intuitively, if the two datasets were very dissimilar, then a classifier would be able to achieve high accuracy. 
Formally, let $\cF$ be class of binary prediction functions mapping each data point (i.e., a row in the tabular dataset) to $\{0,1\}$, then the accuracy of the best $f\in\cF$ on the discriminator task provides a lower bound on the total variation (TV) distance between the two empirical data distributions. 

Hence, we train a predictor $f$ to discriminate between the model-generated data and the census data in order to obtain an empirical lower bound on the distance between the two datasets. Specifically, we concatenate to each model-generated dataset a random sample of $N$ individuals from the ACS census data, and introduce a binary label indicating whether each row of the concatenated dataset was model-generated or not. We then train an XGBoost classifier in this binary prediction task. As an additional point of reference, we also consider the accuracy in discriminating between the census data of any given U.S. state and an equally-sized sample of the ACS data of all other U.S. states. 

\begin{figure}[t!]
\centering
\subfigure{\includegraphics[width=\linewidth]{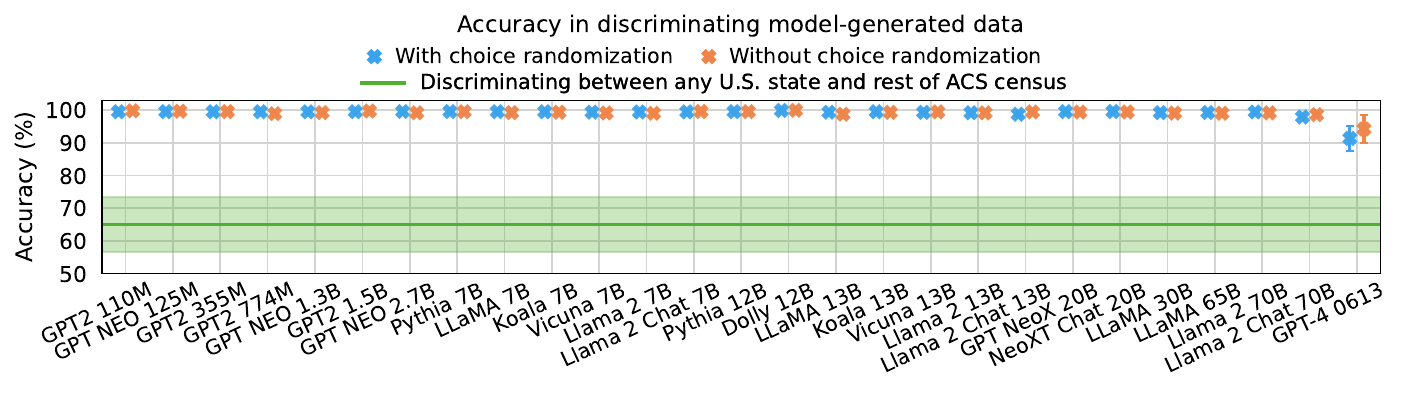}}
\caption{Accuracy of the discriminator test. For all language models, it is possible to discriminate with very high accuracy between the ACS census data and model-generated data, ({\scriptsize\textcolor{myorange}{\faTimes}}) before adjustment and ({\scriptsize\textcolor{myblue}{\faTimes}}) after  adjustment. We contrast this against the accuracy value of discriminating between the ACS data of any given U.S. state and the rest of the ACS census data (\bfseries\textcolor{mygreen}{--}).}
\label{fig:prediction}
\end{figure}

We report mean test accuracy in Figure~\ref{fig:prediction}.
We consider 100 different random seeds. We find that the trained classifiers can differentiate between model-generated data and census data with very high accuracy ($>90\%$) in all cases. Therefore, the empirical distributions corresponding to the model-generated data and the census data have TV distance larger than 0.9. These stark results indicate that data generated by sequentially prompting language models with the ACS survey questionnaire bears little similarity with the data collected by surveying the U.S. population.

\subsection{Contrast with silicon samples}

\citet{argyle2022out}
propose ``silicon sampling'', a methodology to produce synthetic survey respondents using LLMs by conditioning on actual survey respondents. They focus on a subset of 12 questions from the 2016 American National Election Studies (ANES) survey. For every human respondent, they construct a corresponding ``silicon individual'' by querying GPT-3 to predict the ANES respondent's answer to each survey question given the respondent's answers to all other questions.
Their results indicate that, for the 2016 ANES survey, GPT-3 can be a fairly calibrated predictor of an individual's answer to some survey question conditioned on the respondent's answers to all other survey questions.\footnote{\citet{lee2023can} and \citet{sanders2023demonstrations} study imputation tasks similar to those of \citet{argyle2022out}, and find that LLMs are not calibrated predictors for a variety of such tasks.}

However, \citet{argyle2022out} emphasize that important insights can be gained by emulating the survey responses of human populations ``prior to or in the absence of human data''. In this work we have considered precisely the setting where models' responses are obtained in the absence of human data.\footnote{Conceptually, to generate each synthetic individual, we start with a blank survey questionnaire and prompt the LLM to sequentially fill the entire questionnaire. Whereas we only prompt LLMs with their own responses (i.e., to previous survey questions), \citet{argyle2022out} prompt the model only with actual human responses from the 2016 ANES data.}
To investigate how our findings transfer to the ANES, we reproduce the experiments of Section~\ref{sec:generation} using the 2016 ANES survey questionnaire considered by \citet{argyle2022out} and their ``interview-style'' prompt. We apply the discriminator test, and find that the trained classifiers can discriminate between the model-generated data and the ANES data with accuracy $> 99$~\% (see Appendix~\ref{sec:anes}), indicating that models' responses are markedly different to those in the ANES data.

Thus, the fact that models may perform reasonably well at feature imputation tasks (e.g., \emph{predicting} an individual's answer to some question given their answers to \emph{all other questions}) does not imply that models can \emph{generate} synthetic respondents that resemble the responses obtained by surveying human populations. This suggests caution when using LLMs to emulate human populations at present time, in particular in the absence of human data.


\end{document}